\documentclass[11pt]{article}

\usepackage{acl}

\usepackage{times}
\usepackage{latexsym}

\usepackage[T1]{fontenc}

\usepackage[utf8]{inputenc}

\usepackage{microtype}

\usepackage{inconsolata}

\usepackage{amsmath}
\usepackage{appendix}
\usepackage{enumitem} %
\usepackage{hyperref}

\usepackage{algorithm}
\usepackage{algpseudocode}

\makeatletter
\newcounter{phase}[algorithm]
\newlength{\phaserulewidth}
\newcommand{\setphaserulewidth}{\setlength{\phaserulewidth}}
\newcommand{\phase}[1]{%
  \vspace{-1.25ex}
  \Statex\leavevmode\llap{\rule{\dimexpr\labelwidth+\labelsep}{\phaserulewidth}}\rule{\linewidth}{\phaserulewidth}
  \Statex\strut\refstepcounter{phase}\textit{Phase~\thephase~--~#1}%
  \vspace{-1.25ex}\Statex\leavevmode\llap{\rule{\dimexpr\labelwidth+\labelsep}{\phaserulewidth}}\rule{\linewidth}{\phaserulewidth}}
\makeatother

\setphaserulewidth{.7pt}

\usepackage{graphicx}
\usepackage{multirow}
\usepackage{makecell}
\usepackage{booktabs} %
\usepackage{arydshln} %
\usepackage{xcolor} %
\usepackage{soul}
\usepackage{varwidth} %
\usepackage{array} %
\newcolumntype{L}{>{\centering\arraybackslash}m{0.75cm}} %
\newcolumntype{S}{>{\centering\arraybackslash}m{0.6cm}} %
\newcolumntype{M}{>{\centering\arraybackslash}m{0.9cm}} %
\newcolumntype{T}{>{\centering\arraybackslash}m{2mm}} %

\usepackage{tabularx}%

\usepackage{longtable}

\title{Fumbling in Babel:\\ An Investigation into ChatGPT's Language Identification Ability}

\author{Wei-Rui Chen$^{\lambda}$~Ife Adebara$^{\lambda}$~Khai Duy Doan$^{\gamma}$~Qisheng Liao$^{\gamma}$~Muhammad Abdul-Mageed$^{\lambda,\gamma,\psi}$\\ 
  $^{\lambda}$Deep Learning \& Natural Language Processing Group,
  The University of British Columbia\\  $^{\gamma}$Department of Natural Language Processing \& Department of Machine Learning, MBZUAI\\ $^{\psi}$ Invertible AI\\
  \tt \{weirui.chen,ife.adebara,muhammad.mageed\}@ubc.ca, \\
        \tt \{duy.doan,qisheng.liao\}@mbzuai.ac.ae}

\begin{document}
\maketitle
\begin{abstract}
ChatGPT has recently emerged as a powerful NLP tool that can carry out a variety of tasks. However, the range of languages ChatGPT can handle remains largely a mystery. To uncover which languages ChatGPT `knows', we investigate its language identification (LID) abilities. For this purpose, we compile Babel-670, a benchmark comprising $670$ languages representing $24$ language families spoken in five continents. Languages in Babel-670 run the gamut from the very high-resource to the very low-resource. We then study ChatGPT's (both GPT-3.5 and GPT-4) ability to (i) identify  language names and language codes (ii) under zero- and few-shot conditions (iii) with and without provision of a label set. When compared to smaller finetuned LID tools, we find that ChatGPT lags behind. For example, it has poor performance on African languages. We conclude that current large language models would benefit from further development before they can sufficiently serve diverse communities.
\end{abstract}

\section{Introduction}

ChatGPT~\cite{openai2023gpt4} is a large language model (LLM) based on Generative Pre-training (GPT)~\cite{gpt_2020}. It has achieved remarkable success in a wide range of natural language processing (NLP) tasks, including text generation, question answering, and document summarization~\cite{bubeck2023sparks, openai2023gpt4}. The model has been shown to perform well on natural language understanding not only in English, but also Afrikaans, Arabic, Indonesian, Italian, Mandarin Chinese, and several more~\cite{openai2023gpt4}. However, while ChatGPT demonstrates strong language capabilities, it remains unclear what languages it actually ‘knows’. Understanding languages recognized by current Large Language Models (LLMs) empowers the community to set realistic expectations for their application and guides the direction of future development efforts towards particular languages.

\begin{figure}[]
\begin{centering}
\includegraphics[scale=0.105]{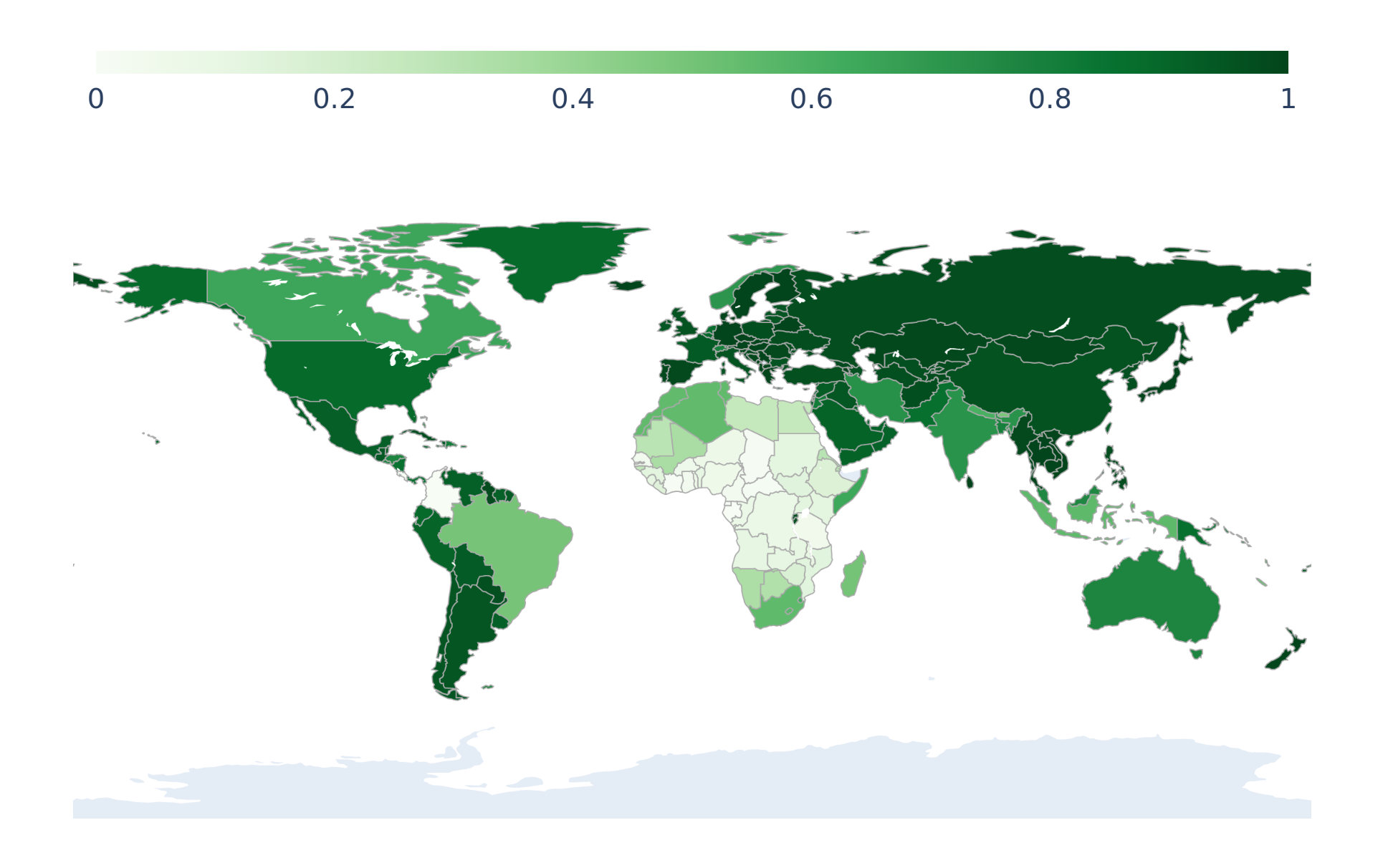}
  \caption{A choropleth map where the intensity indicates the averaged F\textsubscript{1} score of languages spoken in each region. It can be seen that the support of languages has geographical discrepancy, e.g. with African languages being strikingly less supported. The figure is drawn based on the results of one of our experimental settings: (Language Name Prompt [Alias-Dialect-accepting], GPT-4, hard, 0-shot; see Section~\ref{sec:methodology} for more details). A larger map is available in Figure \ref{fig:language_world_map_large} in the Appendix.}
  \label{fig:language_world_map}
 \end{centering}
\end{figure}

Language identification is a fundamental NLP task that plays a critical role in ensuring accurate processing of multilingual data by identifying the language to which a text or speech utterance belongs \cite{tjandraetal2021improved, adebara-etal-2022-afrolid, adebara2023serengeti, burchell-etal-2023-open, madhani-etal-2023-bhasa}. The exponential growth of social media and other digital channels has provided researchers with an abundance of multilingual text. However, \citet{kreutzer-etal-2022-quality} observe datasets being mislabeled with incorrect language and suggest potential risk to downstream applications utilizing these datasets. Hence, LID can be an important step in effectively handling languages and can play a crucial role in the data pipeline of NLP systems~\cite{kreutzer-etal-2022-quality}. For example, \citet{radford2022robust} integrates LID into its pipeline to develop a speech system. LID  also plays a vital role in various NLP applications involving dialects and code-mixed datasets \cite{mageed2020microdialects, thara_etal_2021}. With the emergence of LLMs, there is growing interest in exploring the capabilities of these models for various tasks. Among these, LID, a fundamental NLP task, is notably important to explore.

\begin{figure*}[htp!]
\begin{centering}
\includegraphics[width=0.95\textwidth]{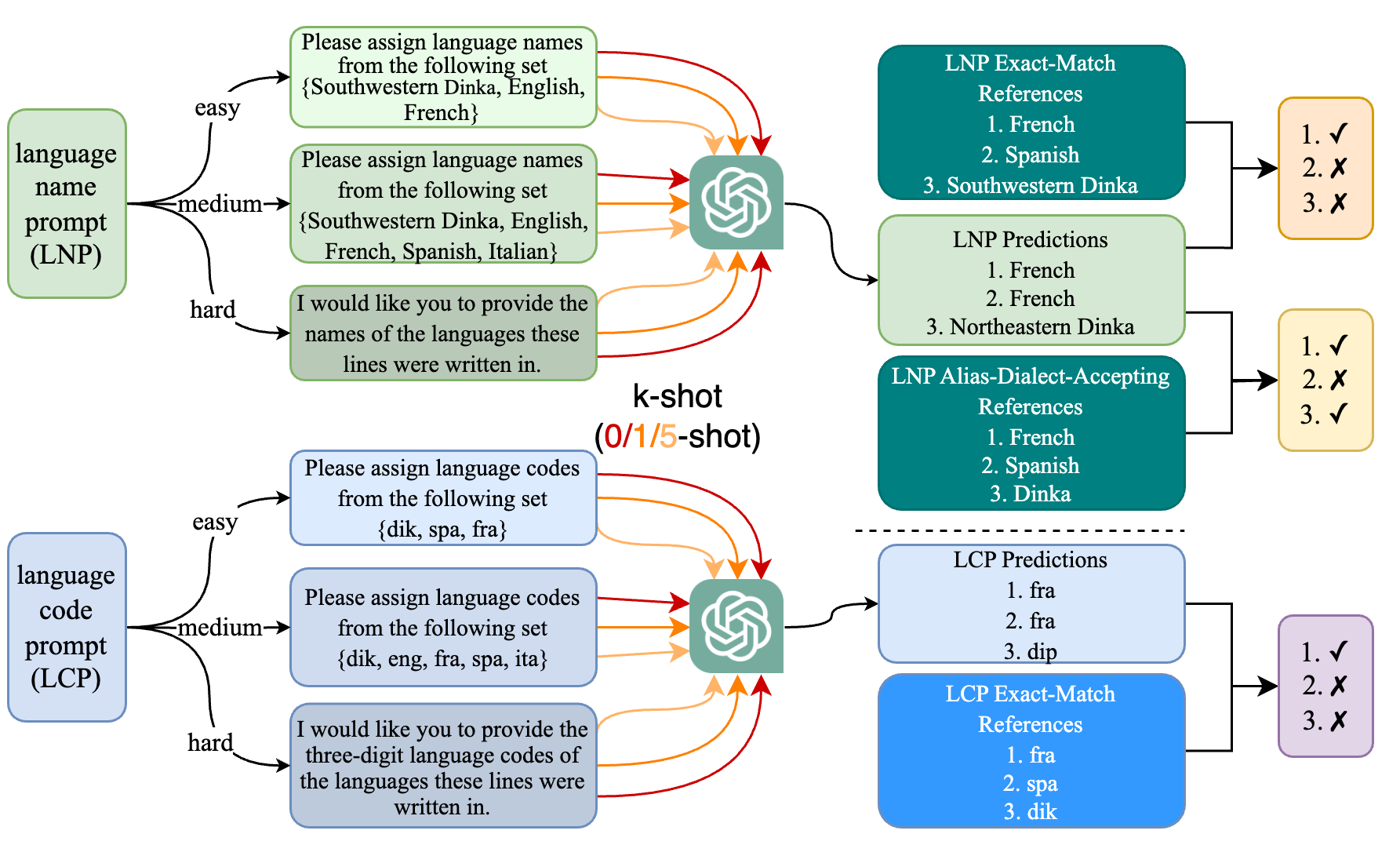}
  \caption{An Overview of different experimental settings with exemplified predictions and test examples in French (fra), Spanish (spa), Southwestern Dinka (dik). Language name prompt (LNP) has both exact-match and alias-dialect-accepting evaluation while language code prompt (LCP) has solely exact-match evaluation. The prediction of third test example (Northeastern Dinka) of LNP is considered incorrect in exact-match evaluation but correct in alias-dialect-accepting evaluation.}
  \label{fig:attraction}
 \end{centering}
\end{figure*}

This paper aims to evaluate the performance of ChatGPT on LID and provides insights into its strengths and limitations. For ChatGPT, we include two backend model checkpoints: \texttt{gpt-3.5-turbo-0613} and \texttt{gpt-4-0613}. Henceforth, we refer to these models as `GPT-3.5' and `GPT-4', respectively. We methodically conduct a series of experiments using diverse text samples from a collection of $670$ languages. We evaluate the model's ability to accurately identify the language of each sample, experimenting with zero- and few-shot settings both with and without the provision of a set of labels. We also carry out a wide range of analyses, including from the perspective of dialectal variation, high-resource and low-resource languages, writing systems (i.e. scripts), and across different geographical locations. We observe that ChatGPT's ability varies remarkably between low-resource and high-resource languages and among different regions as can be seen in Figure~\ref{fig:language_world_map}.

\section{Related Work}\label{sec:related_work}

Traditional approaches to language identification involved rule-based methods~\cite{nakatani2010langdetect}, statistical models~\cite{lui-baldwin-2012-langid}, and handcrafted features such as character combination co-occurrence \cite{van-der-lee-van-den-bosch-2017-exploring, Dongen2017AnalysisAP, Martinc2017PAN2A} and feature smoothing \cite{Jauhiainen2019AutomaticLI}. Recently, deep learning has revolutionized language identification techniques by showing superior performance \cite{jurgens2017incorporating, adebara-etal-2022-afrolid}.

The efficacy of deep learning methods in addressing LID for high-resource languages has generally been established as a resolved issue~\cite{DBLP:journals/corr/abs-2010-14571}. However, the domain of LID for low-resource languages remains significantly under served~\cite{adebara-abdul-mageed-2022-towards}. ChatGPT is a general-purpose language model that is capable of performing a variety of language tasks~\cite{openai2023gpt4}. It has been claimed to have the ability to solve any kind of task without task-specific training or in few-shot settings~\cite{gpt_2020, lin-etal-2022-shot}. In this work, we evaluate the performance of ChatGPT on LID on both low-resource and high-resource languages. To the best of our knowledge, there are no prior works that evaluate ChatGPT on LID. We now introduce Bable-670, our dataset for this work.

\section{Dataset}\label{sec:data}
We curate \textbf{Babel-670}, a dataset for our LID task compiled from three different datasets that cover a total $670$ languages from $24$ language families (shown in Appendix Table~\ref{tab:lang_family_percent}) written in $30$ different scripts (shown in Appendix Table~\ref{tab:script_percent}). A full list of languages included in Babel-670 can be seen in Appendix~\ref{sec:langs_in_babel670}.

\subsection{Data Collection}
Three datasets are curated to build Babel-670. \textbf{AmericasNLP2022} is a dataset that includes five low-resource South American Indigenous languages~\cite{pmlr-v220-ebrahimi22a}. \textbf{AfroLID}~\cite{adebara-etal-2022-afrolid} dataset is manually curated and covers $517$ African languages and language varieties. The dataset is multi-domain and multi-script. \textbf{FLORES-200}~\cite{nllb2022} is a dataset specifically designed for addressing low-resource machine translation, covering $\sim200$ languages. Since there are some languages included in more than one dataset, Bable-670 includes data in the order of AmericasNLP2022, AfroLID, and FLORES-200. That is, if a language is included in more than one dataset, only the data in the dataset of higher order is included. We order these three datasets by their released dates from newer to older as they reflect the chance of the dataset not being included in the training of ChatGPT models. This is to maximize the chance that Babel-670 is unseen during any of its  training phases.

There are no duplicated languages between AmericasNLP2022 and AfroLID datasets, so all languages in these two datasets are included in Babel-670. There are two duplicated languages between AmericasNLP2022 and FLORES-200. There are 46 duplicated languages between AfroLID and Flores-200. For these duplicated languages, the data from AmericasNLP2022 and AfroLID is included. For languages in FLORES-200 which are written in two scripts (e.g. Acehnese as for ace\_Arab and ace\_Latn), we select the script that appears first in alphabetical order. Specifically, for the eight languages in this category, which are Acehnese (\textit{ace}), Modern Standard Arabic (\textit{arb}), Banjar (\textit{bjn}), Kashmiri (\textit{kas}), Central Kanuri (\textit{knc}), Minangkabau (\textit{min}), Tamasheq (\textit{taq}), and Chinese (\textit{zho}), we select Arabic script for the first six languages, Latin for Tamasheq, and Hans for Chinese.

\begin{table}[htp]
\scriptsize
\centering
\begin{tabular}{crrrr}
\toprule
\textbf{Name} & \textbf{\#Langs} & \textbf{Train} & \textbf{Dev} & \textbf{Test} \\
\midrule
AmericasNLP'22 & 5 & 250 & 100 & 75 \\
AfroLID & 517 & 25,850 & 10,340 & 7,755 \\
FLORES-200 & 148 & 7,400 & 2,960 & 2,220 \\
\midrule
Bable-670 & 670 & 33,350 & 13,400 & 10,050\\
\bottomrule
\end{tabular}

\caption{Data splits of our dataset Babel-670.}
\label{tab:dataset}
\end{table}

\subsection{Data Preprocessing}
For each language, there are $50$ training datapoints, $20$ dev datapoints, and $15$ test datapoints extracted from one of the three dataset which is in higher order. Training datapoints are exemplars for few-shot demonstration learning; dev datapoints are used during the development stage when searching for proper prompt templates. All experimental results reported are based on test set. Each prompt contains ten test datapoints which we term a \textit{batch}. A batch would most likely contain test datapoints in different languages since the members of a batch are randomly drawn from test datapoints across all languages without replacement. This design is to avoid having a whole batch of a same language which may affect the performance. However, a batch could occasionally contain datapoints of one language more than once. We keep creating batches until all test datapoints have joined a batch. We choose the batch size to be ten because we try to avoid exceeding the token limit of a API request ($4,096$ for \texttt{gpt-3.5-turbo-0613}, $8,192$ for \texttt{gpt-4-0613}).\footnote{\url{https://platform.openai.com/docs/models} accessed on March, 2024}

For few-shot learning settings, a pool of exemplars is created by incorporating every training data point from all languages. A different number of exemplars (one for 1-shot and five for 5-shot) is randomly drawn from the pool without replacement to join a prompt, along with a batch of test datapoints. For each prompt, there will be no duplication of the exemplars and the test datapoints so ChatGPT will be guided to follow the instruction and the required format without directly being given the answer. %

\section{Methodology}\label{sec:methodology}
To explore ChatGPT's ability to identify languages, we design two major types of prompts: \textbf{language name prompt (LNP)} and \textbf{language code prompt (LCP)}, each encompassing three numbers of shots and three different difficulty levels (See Section~\ref{subsec:diffculty}). An overview of our data pipeline is shown in Figure~\ref{fig:attraction}. LNP asks ChatGPT to predict language names while LCP  asks it to produce three-digit IS0-693 language codes. Although most language identification research use language code as labels, we decide to also prompt ChatGPT to predict language name because (1) it will be a very strong assumption that ChatGPT knows all ISO language codes and (2) we hypothesize ChatGPT is more likely to be fed language names during pretraining. In fact, through analysis, we observe that ChatGPT predicts language names better than language codes given the same set of test examples (see Section~\ref{sec:LNP_vs_LCP}).

\subsection{Prompt Design}
To interact with ChatGPT API effectively, we align our prompt style to the documentations and examples provided by OpenAI (Appendix Table~\ref{tab:openai_documentation_ex}), and adopt temperature of zero to keep the randomness of generation at a low level. A request sent to the API consists of an arbitrary number of messages stored in an ordered json array. There are three types of messages we send to ChatGPT API: \textit{system}, \textit{user}, and \textit{assistant} messages. System message is a high-level instruction to advise the model, typically placed first in the array. User message is where to store what we as users want to communicate with ChatGPT. The assistant message serves two purposes: (1) it records how ChatGPT responds to our user message, and (2) it allows us, as users, to guide ChatGPT by showing the expected response we desire. In the context of few-shot learning, we use a pair of (user, assistant) messages to illustrate the desired behavior to ChatGPT. This pair contains sentences in different languages in the user message and their corresponding gold labels in the assistant message. We structure our few-shot learning examples in a manner similar to ChatGPT's playground example\footnote{\url{https://platform.openai.com/playground} accessed on June, 2023}. This involves using two newline characters (`\textbackslash n') to separate sentences and a colon to indicate that examples follow. Actual test examples are placed in a user message as the last message in the array. To address potential issues, such as irrelevant content or inconsistent outputs, we specify in our prompt that ChatGPT should "never provide anything other than...". This helps ensure precise listings of names or codes and avoids situations where ChatGPT might generate unnecessary information, such as including language codes in LNP or language names in LCP. An example of this is `English (eng)' in a response to our LNP, where the `(eng)' part is unnecessary. For clarity, we provide templates and examples of prompts for different settings in Appendix~\ref{sec:prompt_template_and_prompt_examples}.

We test ChatGPT under three different \textit{k}-shot settings: 0-shot, 1-shot, and 5-shot. For 0-shot, exemplars for demonstration learning is not present and ChatGPT is asked to directly predict the test examples given. For 1-shot and 5-shot, one and five exemplars are given, in a pair of (user, assistant) message, before asking it to predict, respectively. In addition to 'Please answer in ordered listing ...' which is the part in the instruction that specifies response format abstractly, the few shots not only serve as exemplars but also demonstrate to ChatGPT concretely what format we seek for a response.

\subsection{Difficulty Levels}\label{subsec:diffculty}
We design three levels of difficulty to test the ability of ChatGPT: 

\begin{itemize}
\itemsep0em 
  \item Easy: A set of gold labels of the test examples is provided for ChatGPT to choose from.
  \item Medium: Same as easy level but with additional non-gold labels to confuse it.
  \item Hard: No set of labels provided.
\end{itemize}
For the easy and medium levels, a set of language labels (language names for LNP and language codes for LCP) is included in the prompt as a hint for ChatGPT to choose its predictions from. We refer to this set as \textbf{label set}. For the easy level, the size of the label set is equal to the number of unique gold labels for test examples in a batch. For each batch in the medium level, the size of the label set is always 30. That is, in addition to the $\sim10$ unique gold labels, there are $\sim20$ non-gold labels added to the label set\footnote{The number of additional non-gold labels is not always exactly $20$. It will be exactly $20$ if the selected test examples each belongs to a different language. It will be $>20$ if more than one test example belongs to the same language, as the repeated labels will be de-duplicated but the label set is always of size exactly $30$.} to make the task more challenging. We perceive the hard level as the most realistic setting, since it would be rare for the common public to provide a label set for ChatGPT to choose from, instead of directly asking it for an answer. Within the hard level, 0-shot is the use case for average users and few-shot is the use case for researchers and practitioners who may include exemplars in the prompt. On the other hand, medium and easy settings are useful in the sense that they offer an empirical investigation of the text classification ability of ChatGPT when a set of labels is given.

\subsection{Postprocessing and Evaluation}\label{sec:evaluation}
We report our results in accuracy and F\textsubscript{1} score for every experimental setting. Before we evaluate, we postprocess the output of ChatGPT as it is in textual format and cannot be compared with our labels directly. 

\noindent \textbf{Postprocessing.} As ChatGPT is a generative model and produces output in textual format, it takes postprcoessing of the output to extract predictions which can be later used for evaluation. Since we ask ChatGPT to answer in ordered listing in our prompt, each pair of number and the following string are extracted where the latter is taken as prediction. We expect the number of extracted pairs to be identical to that of the batch size (i.e. 10). If not, a `None' will be inserted along with the missing number.\footnote{After an inspection, we found that all test examples are given a prediction. Therefore, no test example is assigned `None'.} Having pairs ready, we extract the content that follows each number. For LCP, a typical pair is such as `10. kmb'. We extract the first occurrence of alphabetical substring (`kmb') coming after the number and ignore the rest. This is to only extract language code itself as occasionally ChatGPT produces language name alongside language code (e.g. `mkd [Macedonian]) despite being explicitly instructed not to do so. For LNP, we extract the alphabetical sequence that follows a digit. Unlike LCP, we include the whole following alphabetical sequence so language names consisting of more than one word can be extracted successfully, e.g. `Egyptian Arabic'. 

\noindent \textbf{Evaluation.} For LCP, only when the  prediction and gold label is identical will it be considered a hit; otherwise, a miss. We name this \textbf{exact-match} evaluation. For LNP, we report results based on not only exact-match but also \textbf{alias-dialect-accepting} evaluation. We propose this because there is a fundamental difference between classification model and generative model. The prediction of a classification model always falls within a preset closed space (i.e. the classes). However, the prediction of ChatGPT does not guarantee that. Additionally, unlike ISO language code system which in general follows the principle of having one language code per language, the fact that a language can have multiple names makes LNP evaluation more challenging. Given the discrepancy, an exact-match evaluation for LNP may not reflect the true ability of a generative model because of potentially considerable number of false negatives. Alias-dialect-accepting evaluation is a fuzzy matching strategy that reduces the number of false negatives. For example, under exact-match evaluation, it will be considered a miss if the model predicts `Español' when the label is `Spanish', even though they are referring to the same language entity and can be an alias for each other. Another example is if it predicts `Northeastern Dinka' when the label is `Southwestern Dinka'; they belong to the same language group Dinka as dialects. We propose alias-dialect-accepting evaluation to address these two issues. Under this setting, if ChatGPT's prediction is an alias of the ground truth or if prediction and ground truth belong to a same language group, it will be counted as a hit. Implementation details of the alias-dialect-accepting evaluation are in Appendix~\ref{sec:ada_evaluation}.

\begin{table*}[h]
\footnotesize

\centering
\begin{tabular}{cccccccccccccccc}
\toprule
Level & \# & \multicolumn{7}{c}{GPT-3.5-turbo-0613} & \multicolumn{7}{c}{GPT-4-0613} \\

\cmidrule(lr){3-8} 
\cmidrule(lr){9-14} 

& 
& \multicolumn{2}{c}{LNP (exact)}  & \multicolumn{2}{c}{LNP (ADA)} & \multicolumn{2}{c}{LCP (exact)} & \multicolumn{2}{c}{LNP (exact)} & \multicolumn{2}{c}{LNP (ADA)} & \multicolumn{2}{c}{LCP (exact)} \\

\cmidrule(lr){3-4}
\cmidrule(lr){5-6}
\cmidrule(lr){7-8}
\cmidrule(lr){9-10}
\cmidrule(lr){11-12}
\cmidrule(lr){13-14}
&
& ACC & $F_1$  & ACC & $F_1$ & ACC & $F_1$ & ACC & $F_1$  & ACC & $F_1$ & ACC & $F_1$ \\
\midrule
\multirow{3}{*}{easy} & 0 & 24.86 & 28.28 & 24.90 & 28.14 & 14.63 & 17.63 & 65.82 & 65.36 & 65.97 & 65.92 & 47.16 & 46.29 \\
& 1 & 28.13 & 31.18 & 28.23 & 31.12 & 27.32 & 27.51 & 65.19 & 65.70 & 65.29 & 66.16 & 45.88 & 45.96 \\
& 5 & 33.82 & 34.46 & 33.96 & 34.97 & 27.37 & 27.17 & 68.29 & 68.05 & 68.42 & 68.69 & 47.95 & 47.31 \\
\midrule
\multirow{3}{*}{medium} & 0 & 19.81 & 21.09 & 20.26 & 21.62 & 10.27 & 12.30 & 48.66 & 45.39 & 49.08 & 46.28 & 37.79 & 34.24 \\
& 1 & 25.30 & 22.63 & 26.32 & 23.58 & 22.88 & 19.60 & 48.02 & 45.57 & 48.40 & 46.22 & 38.19 & 34.64 \\
& 5 & 26.22 & 23.02 & 26.94 & 24.08 & 22.76 & 19.64 & 50.49 & 47.63 & 50.89 & 48.32 & 39.39 & 35.62 \\
\midrule
\multirow{3}{*}{hard} & 0 & 12.70 & 12.17 & 17.39 & 16.36 & 2.47 & 3.82 & 20.02 & 17.80 & 28.32 & 24.16 & 21.47 & 18.93 \\
& 1 & 16.05 & 13.81 & 23.82 & 19.36 & 16.66 & 14.34 & 20.40 & 18.26 & 28.58 & 24.76 & 22.10 & 19.72 \\
& 5 & 17.12 & 14.65 & 25.36 & 20.25 & 17.71 & 15.09 & 20.94 & 18.97 & 28.79 & 25.31 & 22.52 & 20.26 \\

\bottomrule
\end{tabular}
\caption{Accuracy values (\%) and macro-averaged $F_1$ scores (\%) of different experimental settings for GPT-3.5 and GPT-4. \textbf{Level}: difficulty level, \textbf{\#}: number of shot(s), \textbf{LNP}: language name prompt, \textbf{LCP}: language code prompt, \textbf{exact}: exact-match evaluation, \textbf{ADA}: alias-dialect-accepting (ADA) evaluation. ADA is only applicable for LNP. }
\label{tab:results}
\end{table*}

\section{Results and Analysis}\label{sec:results}

\subsection{Comparison of Different Settings}
This section includes the experimental results for all settings (see Table~\ref{tab:results}) and the analyses conducted to compare each pair of different settings. 
\subsubsection{LNP vs. LCP}\label{sec:LNP_vs_LCP}
For easy and medium difficulty levels, LNP always has better performance than LCP across all settings. For hard level, the performance of LNP with exact-match evaluation is modestly inferior to LCP while LNP with alias-dialect-accepting is significantly better than LCP in all settings. For hard level where there is no label set provided, LNP with exact-match has a fundamental limitation as discussed in Section~\ref{sec:evaluation} which may contribute to it slightly underperforming LCP. We argue that when it comes to language identification, ChatGPT knows language names better than language codes. That is, for the same given piece of text, ChatGPT is more likely to correctly identify its language if it is asked to produce language name rather than language code. We speculate this is the case since language names are much more likely to occur than language codes in the pretraining data of ChatGPT.

Furthermore, we observe that GPT-3.5 lacks a robust understanding of the concept of language codes. It faces challenges when tasked with identifying language codes in the absence of exemplars or label sets, correctly identifying only $2.47\%$ of all test examples under the hard level and 0-shot conditions. However, its performance sees an improvement of  $\sim574\%$ when one exemplar is provided. This pattern of profound improvement going from 0-shot to 1-shot persists in medium and easy settings. In contrast, GPT-4 exhibits a more proficient understanding of language codes. The presence or absence of exemplars has a lesser impact on performance compared to GPT-3.5.

\subsubsection{Difficulty levels}
The difference between difficulty levels is the provision and size of a label set in the prompt. Since there are around $7,000$ human languages, performing language identification without a label set is similar to performing a $7000$-class text classification which can be challenging. The provision of a label set limits the range of output values and therefore improves the manageability of the task. Moreover, the smaller the provided label set, the less challenging the task. We observe a significant performance difference in the rank $\text{easy} > \text{medium} > \text{hard}$. For (LNP [alias-dialect-accepting], GPT-4, 0-shot) setting, the accuracy and F\textsubscript{1} score for easy level are $34.4\%$ and $42.4\%$ higher than those for medium level, respectively. The accuracy and F\textsubscript{1} score for medium level are $73.3\%$ and $91.6\%$ higher than those for hard level, respectively. Settings under GPT-3.5 and for LCP have show similar performance patterns. However, if ChatGPT truly identifies those languages, it should perform similarly regardless of the provision and size of the label set.

We argue that significant performance disparities exist between different difficulty levels due to two primary reasons: \textbf{(1)} In cases where ChatGPT has no prior knowledge about the language of a test example, providing a label set increases the likelihood of correct guessing. This is because, when a smaller label set is available, ChatGPT can randomly assign a label from the set, resulting in a higher probability of a fortunate correct guess. Probabilistically, the average number of successful classifications for a set of examples randomly assigned to $10$ classes (easy level) is much higher than that for $30$ (medium level). \textbf{(2)} For test examples where ChatGPT possesses some knowledge but lacks confidence in determining the language due to factors such as code-switched text, brevity, or closely related languages with shared vocabulary and linguistic characteristics, the provision of a label set boosts confidence by eliminating numerous potential candidates. Smaller label sets reduce the number of candidates to consider. Given this analysis, we posit that the $\sim70\%$ accuracy achieved by ChatGPT under the setting (LNP[alias-dialect-accepting], GPT-4, easy, 5-shot) may present an overly optimistic estimation of its capabilities.

\subsubsection{GPT-3.5 vs. GPT-4}\label{sec:gpt3.5_vs_gpt4}

As anticipated, GPT-4 consistently exhibits stronger performance than GPT-3.5 across all settings. Particularly in easy and medium difficulty levels, GPT-4 manages to double the performance of GPT-3.5 in numerous scenarios, suggesting a potentially superior natural language understanding ability. However, in the hard level, GPT-4 outperforms GPT-3.5 by smaller margins. As discussed later in section~\ref{sec:low_precision}, we argue that the hard level best reflects the true language identification capability of a model. Hence, we speculate that the narrow performance gap is likely due to GPT-4's slightly broader range of supported languages compared to GPT-3.5. In other words, if the number of supported languages were to increase significantly, we would expect a larger performance gap in the hard level. 

Furthermore, GPT-4 proves to be a superior zero-shot learner compared to GPT-3.5, whose zero-shot performance is limited in comparison. GPT-3.5's improvement from 0-shot to 1-shot is much more substantial than that from 1-shot to 5-shot, indicating the necessity of including at least one exemplar for GPT-3.5. Conversely, for GPT-4, performance remains consistent across different numbers of shots, highlighting its enhanced capability to perform tasks without exemplars.

\begin{table}[htp]
\tiny
\centering

\begin{tabular}{lccccccc}
\toprule
\multirow{2}{*}{Tool} & \multirow{2}{*}{\#lang} & \multirow{2}{*}{Acc} & \multirow{2}{*}{F\textsubscript{1}} & \multicolumn{2}{c}{GPT-3.5} & \multicolumn{2}{c}{GPT-4} \\
\cmidrule(lr){5-6}
\cmidrule(lr){7-8}

\multicolumn{1}{c}{}   &        &        &        & Acc      & F\textsubscript{1}       & Acc       & F\textsubscript{1}   \\  
\midrule
AfroLID&      517      &      92.90         & 89.04   &0.55         &   0.82     &  7.72      &  4.79   \\ 
CLD2 & 66 & 96.03 &91.22&15.05&8.45&95.45&83.81  \\
CLD3 & 83 & 96.02&89.53&14.86&8.58&93.65&72.07  \\
FastText &101 & 83.77&74.02&12.61&7.67&88.05&64.25 \\
Franc &216 &81.05&66.28 &7.08&5.32&56.87&29.73 \\
LangDetect & 48 & 99.03&99.01&15.83&8.63&97.78&92.69  \\
Langid.py  & 78 & 92.39&88.80&14.61&8.17&91.03&71.88  \\ 
\bottomrule
\end{tabular}

\caption{Comparison of accuracy and macro-averaged F\textsubscript{1} score (\%) to other language identification tools on languages supported by the tool and are included in Babel-670. For ChatGPT models, they are of setting (LCP, hard, 0-shot). GPT-3.5 performs poorly partially because of its inability of understanding the instruction under 0-shot (discussion can be seen in section~\ref{sec:gpt3.5_vs_gpt4}).}
\label{tab:comparison_other_tools}
\end{table}

\subsection{Comparison to Other Tools}\label{sec:comparison_to_other_tools}
We conduct a performance comparison between GPT-3.5, GPT-4, and other language identification tools. Specifically, we compare with AfroLID~\cite{adebara-etal-2022-afrolid}, CLD2, CLD3~\cite{salcianu2018compact}, FastText~\cite{joulin2016bag}, Franc,  LangDetect~\cite{nakatani2010langdetect}, and Langid.py~\cite{lui-baldwin-2012-langid}.\footnote{Detailed information (version, license and URL) of these tools are included in Appendix Table~\ref{tab:artifacts_info}} For each tool, we evaluate only the languages that are both supported by the tool and are included in Babel-670. We only include the setting (LCP, hard, 0-shot) to have a fair comparison to other tools as the tools are all (1) language code-based (2) do not allow in-context learning (3) do not allow label set.

As shown in Table~\ref{tab:comparison_other_tools}, all tools outperform GPT-3.5 and GPT-4, except FastText where GPT-4 demonstrates better performance in accuracy. Our assessment reveals the lowest performance exhibited by both GPT-3.5 and GPT-4 is in the context of African languages when comparing with AfroLID, which includes only African languages. GPT-3.5 has an extremely low F\textsubscript{1} score at  $0.82\%$ while GPT-4 has a better but still limited performance of F\textsubscript{1} score at $4.79\%$.

\begin{figure*}[]
\begin{centering}
\includegraphics[scale=0.323]{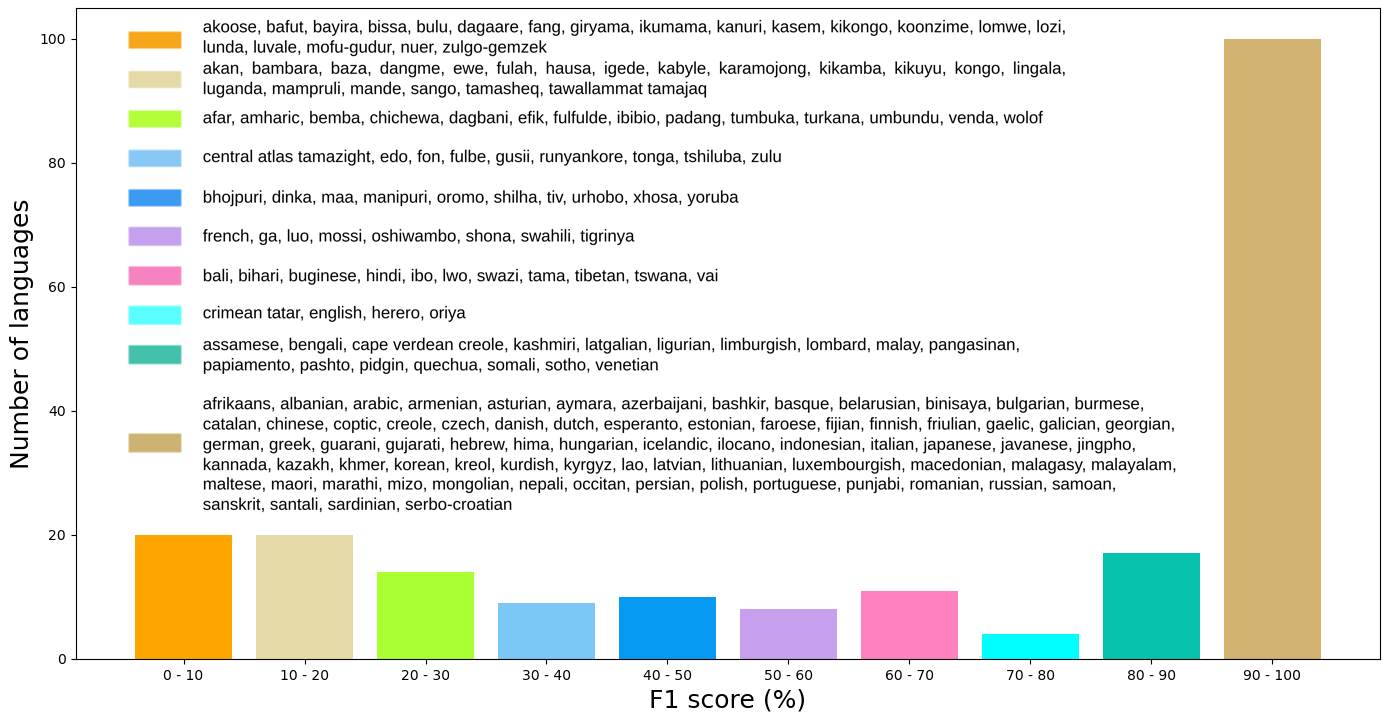}
  \caption{Languages with different ranges of F1 scores ($\%$). 382 languages with zero $F_1$ score are not included in this figure but are reported in Appendix Table~\ref{tab:lang_names_zero_f1_LNP_dialect_accepting}. It shows a M-shape bimodal  distribution where both extremes, zero F\textsubscript{1} score for $382$ languages and $>90\%$ F\textsubscript{1} score for $100$ languages, take up most languages ($\sim 500$ languages). This is of setting (LNP [alias-dialect-accepting], GPT-4, hard, 0-shot). }
  \label{fig:f1_bar_plot_LNP_dialect_accepting_hard_0shot_gpt4}
 \end{centering}
\end{figure*}

\subsection{Error Analysis}\label{sec:low_precision}
To perform the error analysis, we analyze the (LNP [alias-dialect-accepting], GPT-4, hard, 0-shot) setting. We use this setting for three main reasons: \textbf{(1)} Alias-dialect-accepting evaluation provides a more accurate measure of a generative model's capabilities, as discussed in Section~\ref{sec:evaluation}. \textbf{(2)} Hard level best reflects the actual LID ability of a model since there is little chance for a lucky guess as label set is not provided. \textbf{(3)} Hard level with 0-shot best reflects practical usage scenarios for the general public. In many scenarios, users may not have access to demonstration learning or the ability to provide a specific label set when attempting to identify a piece of text. In Figure \ref{fig:f1_bar_plot_LNP_dialect_accepting_hard_0shot_gpt4}, we show the F\textsubscript{1} scores of the languages that achieve an F\textsubscript{1} greater than zero.

We find that languages achieving the highest F\textsubscript{1} scores are those primarily categorized as ``rising stars", ``underdogs", and ``winners", as defined by Joshi et al. \cite{joshi-etal-2020-state}. \texttt{Rising stars} possess a substantial amount of unlabeled data but face constraints due to limited labeled data. In contrast, \texttt{underdogs} have a comparable quantity of unlabeled data to \texttt{winners}, but they have relatively fewer labeled examples. \texttt{Winners} which represent approximately seven languages globally, benefit from abundant resources for NLP tasks.

However, it is noteworthy that some languages classified as \texttt{Winners} exhibit unexpectedly low performance. For example, English and French have F\textsubscript{1} scores of $76.92\%$ and $56.23\%$, respectively. In the case of English, while all English examples in our test data are correctly labeled, numerous examples of other languages are incorrectly classified as English, including English-based creoles \cite{adebara-etal-2022-afrolid} like Nigerian Pidgin and Cameroonian Pidgin, as well as languages like Somali, Swahili, Harari, and Kinyarwanda, which feature some code-mixing in their data. Similarly, all French examples are correctly labeled as French, but several other languages are mistakenly classified as French. These misclassified languages are those spoken in Francophone Africa, which exhibit some degree of code-mixing with French \cite{codeswitching_2014}.

Furthermore, we have observed that many African languages with diacritics in their orthographies are incorrectly predicted as Yoruba. For example, Gokana, Igala, Keliko, Yala, Igede, and several others are inaccurately labeled as Yoruba. Additionally, languages that share a few vocabulary items with Yoruba are also misclassified as Yoruba. Languages like Oshiwambo, Mogofin, and Rigwe fall under this category. This suggests that the models only adequately support a limited number of African languages with diacritics in their orthographies. Intriguingly, we have identified cases of very low-resource languages, falling into the categories of \texttt{left-behinds} and \texttt{scrapping-bys} achieving unexpectedly high F\textsubscript{1} scores. Languages like Gaelic, Guarani, Jingpho, and Kurdish fall into this category. It is plausible that the data used in our test set may have been included in the training data for the GPT models, resulting in these high F\textsubscript{1} scores. We make this assumption because \texttt{left-behinds} and \texttt{scrapping-bys} languages have exceptionally limited data for NLP work.

\subsection{Geographical Analysis} 
We conduct an analysis from a geographical perspective, visualizing our model performance with a choropleth map as can be seen in Figure~\ref{fig:language_world_map} (a large version of the same figure can be seen at Appendix Figure~\ref{fig:language_world_map_large}). The map is drawn based on the results of (LNP
[alias-dialect-accepting], GPT-4, hard, 0-shot) setting.  We consult Ethnologue\footnote{https://www.ethnologue.com/} to retrieve the region(s) in which each of our Babel-670 languages are primarily spoken. Certain regions remain uncolored either because there are no languages  spoken in the region (e.g. Antarctica) or the languages spoken there are not covered in the Babel-670 dataset. It is important to highlight the limitation that Babel-670 does not cover all human languages and therefore the visualization can only reflect the state of languages included in our Babel-670 dataset. Notably, Africa demonstrates the lightest colors, aligning with the findings of the LCP setting discussed in section~\ref{sec:comparison_to_other_tools}. This highlights ChatGPT's limited support for African languages, underscoring the importance of inclusion of languages with less digital resources and representation. It also indicates that ChatGPT has not reached the state of serving diverse communities.

\subsection{Script-Wise Analysis}
Previous research suggests that languages with unique writing systems are more easily distinguishable by language identification models \cite{jauhiainen-etal-2017-evaluation, adebara-etal-2022-afrolid,adebara2023improving}. In our study, which spans $30$ scripts, we investigate this observation further. Our analysis shows that languages utilizing distinct scripts generally achieve higher F\textsubscript{1} score, as demonstrated in Table~\ref{tab:script_num_avgF1}. For example, scripts such as Japanese and Hangul, corresponding to Japanese and Korean languages respectively, attain perfect F\textsubscript{1} scores of $100\%$. Conversely, scripts shared by a larger number of languages, including Arabic, Devanagari, and Latin, are associated with lower F\textsubscript{1} scores. The Latin script, used by the most extensive array of languages in our study, notably averages an F\textsubscript{1} score of $17.64\%$.

\begin{table}[]
\scriptsize
\centering
\begin{tabular}{lrrlrr}
\toprule
\textbf{Script}     & \textbf{\#} & \textbf{Avg F\textsubscript{1}} & \textbf{Script}    & \textbf{\#}  & \textbf{Avg F\textsubscript{1}} \\
\midrule
Arabic     & 18 & 54.81 & Hebrew    & 2   & 95.45 \\
Armenian   & 1  & 100   & Japanese  & 1   & 100   \\
Bengali    & 3  & 69.12 & Kannada   & 1   & 100   \\
Burmese    & 2  & 100   & Khmer     & 1   & 100   \\
Coptic     & 1  & 96.77 & Laoo      & 1   & 100   \\
Cyrillic   & 11 & 98.78 & Latin     & 581 & 17.64 \\
Devanagari & 8  & 64.34 & Malayalam & 1   & 100   \\
Ethiopic   & 6  & 13.98 & Odia      & 1   & 71.42 \\
Georgian   & 1  & 100   & Ol Chiki  & 1   & 100   \\
Greek      & 1  & 100   & Sinhala   & 1   & 100   \\
Gujarati   & 1  & 100   & Tamil     & 1   & 100   \\
Gurmukhi   & 1  & 100   & Telugu    & 1   & 96.77 \\
Hangul     & 1  & 96.77 & Thai      & 1   & 100   \\
Hans       & 1  & 95.24 & Tibetan   & 2   & 33.33 \\
Hant       & 1  & 95.24 & Vai       & 1   & 63.63 \\
\bottomrule
\end{tabular}
\caption{30 scripts and the number (\#) and average F\textsubscript{1} scores (\%) of languages written in each of these scripts in Babel-670.}\label{tab:script_num_avgF1}
\end{table}

Building upon this observation, we propose a hypothesis suggesting that scripts employed by fewer languages may be more easily identifiable by a language identification system, leading to higher F\textsubscript{1} scores owing to their inherent distinctiveness. Specifically, we posit a negative correlation between the number of languages utilizing a particular script and the average F\textsubscript{1} score of those languages. To validate this hypothesis, we perform correlation analysis on the 30 scripts employed by the languages in Babel-670, with the setting (LNP [alias-dialect-accepting], GPT-4, hard, 0-shot).

The correlation analysis shows a significant negative correlation across all three correlation methods: Pearson's $r$ (-0.52), Kendall's $\tau$ (-0.54), and Spearman's rank (-0.63), all having \textit{p}-value $<.01$. This confirms our hypothesis.

In contrast to Table~\ref{tab:script_percent}, some scripts in Table~\ref{tab:script_num_avgF1} have a smaller number of languages. This is because we exclude languages which belong to a language group having more than one script, after categorizing these by our proposed alias-dialect-accepting evaluation method (see Appendix~\ref{sec:ada_evaluation} for details). For example, the language group Serbo-Croatian and its member languages (Serbian (\textit{srp}), Bosnian (\textit{bos}), Croatian (\textit{hrv})) are excluded because Serbian utilizes Cyrillic script while Bosnian and Croatian use Latin script. Including the F\textsubscript{1} score of Serbo-Croatian for computation of average F\textsubscript{1} for Cyrillic script is biased as the group Serbo-Croatian includes languages that use Latin script. For a similar reason, the F\textsubscript{1} of Serbo-Croatian is not included in the computation of average F\textsubscript{1} for Latin script. 

To retrieve the script utilized by each language, we consult Ethnologue and the script information described in FLORES-200 webpage.\footnote{\url{https://github.com/facebookresearch/flores/blob/main/flores200/README.md} accessed on March, 2024.} If it is not available in these two sources, we manually inspect the script in the data.

\section{Conclusion}\label{sec:conclusion}

To investigate ChatGPT's ability to identify human languages, we curate \textit{Babel-670}, which is a dataset that covers $670$ languages spoken in five continents, belonging to 24 language families and are written in $30$ different scripts. We prompt two versions of ChatGPT to produce language names and language codes, each with a different number of exemplars with and without provision of a label set. We conduct comprehensive analyses focusing on errors, geographic distribution, and script variations on the results retrieved with our proposed novel evaluation method that takes language aliases and dialects into consideration. We find that ChatGPT has an uneven ability at identifying languages. The model is able to identify one hundred languages at $>90\%$ F\textsubscript{1} score but has entirely deficient knowledge for another 382 languages (where it achieves a zero F\textsubscript{1} score). Geographically, among the five continents, African languages have the least support by ChatGPT. The investigation demonstrates that ChatGPT is still a considerable distance away from serving wide and diverse communities adequately.

\section*{Limitations}
We identify the following limitaions for this work:
\begin{itemize}
    \item \textbf{Representativeness of World Languages} Our goal is to encompass a broad spectrum of linguistic diversity by incorporating the Babel-670 collection, which represents a significant portion of global languages. Despite our efforts, it is important to acknowledge the inherent limitations of this approach, as the vast linguistic landscape of approximately 7,000 known human languages extends beyond our current scope. Therefore, the analyses covered in our work should therefore be interpreted as an illustration of the capabilities over the languages in Babel-670 dataset, rather than a comprehensive global linguistic representation. Also, ChatGPT's proficiency in language identification of one language does not necessarily translate to comparable performance in more complex downstream tasks in the same language.
    
    \item \textbf{Ethnologue Coverage.} For the alias-dialect-accepting evaluation, we curate a set of language names from Ethnologue and Python package \verb|langcodes|. It is important to mention that there are seven codes not recognized by Ethnologue: ngo, nob, fat, ber, ajp, nno, and twi. Therefore, these seven languages have a single language name, unlike many other languages having multiple names.

    \item Creating the choropleth map involves utilizing data sourced from Ethnologue, which introduces several unique challenges. These challenges include:
    \begin{enumerate}
        \item \textbf{Data Updates.} Ethnologue regularly updates its information on their website. Consequently, we cannot ensure that the data used in this work represent the most recent updates from Ethnologue as we access it at  different points in time.
        \item \textbf{Divergence From Other Sources.} The information concerning languages, dialects, and their associated countries may differ from that found in other sources. This divergence may not always accurately reflect the actual linguistic landscape.
        \item \textbf{Equal Weights for All Languages.} In the process of constructing the map, we assign equal weights to all languages spoken in a certain region, regardless of the number of speakers. For instance, languages like English and French, which have speaker populations of approximately 75\% and 23\%,\footnote{According to \url{https://www.canada.ca/en/canadian-heritage/services/official-languages-bilingualism/publications/statistics.html}, accessed on Nov 5, 2023} respectively, in Canada, receive the same weight. This approach can result in the map not fully reflecting the specific support that ChatGPT offers to different languages in various regions from the perspective of population.
    \end{enumerate}

\end{itemize}

\section*{Ethics Statement}
We would like to make the following ethics-related statements about our work:

\begin{itemize}
\item Our research at its core aims at identifying limitations of current technologies and motivating expansion of their coverage. We perceive this objective as aligning with efforts to improve equity and diversity in AI, an important undertaking necessary for the wide populations of technology users.  

\item Another ethics-pertaining aspect is our datasets: The datasets we use are derived from previous research and are all collected from the public sphere. For these reasons, we do not have serious concerns about use of these for our research. 

\item As we do not develop models in this work, we do not have concerns related to model use. 
\end{itemize}

\section*{Acknowledgements}\label{sec:acknow}
We acknowledge support from Canada Research Chairs (CRC), the Natural Sciences and Engineering Research Council of Canada (NSERC; RGPIN-2018-04267), the Social Sciences and Humanities Research Council of Canada (SSHRC; 895-2020-1004; 895-2021-1008), Canadian Foundation for Innovation (CFI; 37771), Digital Research Alliance of Canada,\footnote{\href{https://alliancecan.ca}{https://alliancecan.ca}} and UBC ARC-Sockeye.\footnote{\href{https://arc.ubc.ca/ubc-arc-sockeye}{https://arc.ubc.ca/ubc-arc-sockeye}}

\bibliography{custom}

\appendix
\clearpage %
\onecolumn
\appendixpage
\addappheadtotoc
\counterwithin{figure}{section}
\counterwithin{table}{section}
There are four sections in the appendix:
\begin{itemize}
    \item Appendix~\ref{sec:miscellaneous} includes tables and figures that are referred in the main content.
    
    \item Appendix~\ref{sec:prompt_template_and_prompt_examples} covers the prompt templates for both language name prompt (LNP) and language code prompt (LCP) under different difficulty levels and number of shots. Example prompts are also given.
    
   \item Appendix~\ref{sec:ada_evaluation} includes implementation details of proposed alias-dialect-accepting evaluation method.
   \item Appendix~\ref{sec:langs_in_babel670} includes a full list of languages included in Babel-670.
\end{itemize}

\section{Miscellaneous}\label{sec:miscellaneous}

\begin{table}[h!]
\scriptsize
\centering
\begin{tabular}{lrrlrr}
\toprule
Family     & \#                        & \%   & Family    & \#  & \%    \\
\midrule
Afro-Asiatic         & 72 & 10.75 & Koreanic         & 1   & 0.15  \\
Austro-Asiatic       & 3  & 0.45  & Kra-Dai          & 3   & 0.45  \\
Austronesian         & 21 & 3.13  & Language Isolate & 1   & 0.15  \\
Aymaran              & 1  & 0.15  & Mongolic         & 1   & 0.15  \\
Chibchan             & 1  & 0.15  & Niger Congo      & 386 & 57.61 \\
Creole               & 12 & 1.79  & Nilo-Saharan     & 57  & 8.51  \\
Constructed Language & 1  & 0.15  & Quechuan         & 1   & 0.15  \\
Dravidian            & 4  & 0.60   & Sino-Tibetan     & 9   & 1.34  \\
Indo-European        & 74 & 11.04 & Tucanoan         & 2   & 0.30  \\
Japonic              & 1  & 0.15  & Tupian           & 1   & 0.15  \\
Kartvelian           & 1  & 0.15  & Turkic           & 11  & 1.64  \\
Khoe-Kwadi           & 3  & 0.45  & Uralic           & 3   & 0.45  \\
\bottomrule
\end{tabular}
\caption{24 Language families and the number (\#) and proportion (\%) of languages within each language family in Babel-670.}
\label{tab:lang_family_percent}
\end{table}
\begin{table}[h!]
\scriptsize
\centering
\begin{tabular}{lrrlrr}
\toprule
Script     & \#                        & \%   & Script    & \#  & \%    \\
\midrule
Arabic     & 23 & 3.43 & Hebrew    & 2   & 0.30  \\
Armenian   & 1  & 0.15 & Japanese  & 1   & 0.15  \\
Bengali    & 3  & 0.45 & Kannada   & 1   & 0.15  \\
Burmese    & 2  & 0.30 & Khmer     & 1   & 0.15  \\
Coptic     & 1  & 0.15 & Laoo      & 1   & 0.15  \\
Cyrillic   & 12 & 1.79 & Latin     & 589 & 87.91 \\
Devanagari & 10 & 1.49 & Malayalam & 1   & 0.15  \\
Ethiopic   & 6  & 0.9  & Odia      & 1   & 0.15  \\
Georgian   & 1  & 0.15 & Ol Chiki  & 1   & 0.15  \\
Greek      & 1  & 0.15 & Sinhala   & 1   & 0.15  \\
Gujarati   & 1  & 0.15 & Tamil     & 1   & 0.15  \\
Gurmukhi   & 1  & 0.15 & Telugu    & 1   & 0.15  \\
Hangle     & 1  & 0.15 & Thai      & 1   & 0.15  \\
Hans       & 1  & 0.15 & Tibetan   & 2   & 0.30  \\
Hant       & 1  & 0.15 & Vai       & 1   & 0.15  \\
\bottomrule
\end{tabular}
\caption{30 scripts and the number (\#) and proportion (\%) of languages written in each of these scripts in Babel-670.}
\label{tab:script_percent}
\end{table}
\begin{table*}[h!]
\footnotesize
\centering
\begin{tabular}{p{15cm}}
\toprule
\texttt{messages=[
        \newline
        \{"role": "system", "content": "You are a helpful assistant."\},
        \newline
        \{"role": "user", "content": "Who won the world series in 2020?"\},
        \newline
        \{"role": "assistant", "content": "The Los Angeles Dodgers won the World Series in 2020."\},
        \newline
        \{"role": "user", "content": "Where was it played?"\}
        \newline
    ]
}
\\
\bottomrule
\end{tabular}
\caption{The example request shown in OpenAI documentation for ChatGPT API. It is a json array with a system message, a pair of (user, assistant) messages, and a user message at last position which contains real question to ask ChatGPT. The pair of (user, assistant) messages is for demonstration learning, showing ChatGPT how the request sender wishes the conversation to be like. The documentation is accessed on July 12, 2023 at \url{https://platform.openai.com/docs/guides/gpt/chat-completions-api}. }
\label{tab:openai_documentation_ex}
\end{table*}

\begin{figure*}[h!]
\begin{centering}
\includegraphics[scale=0.20]{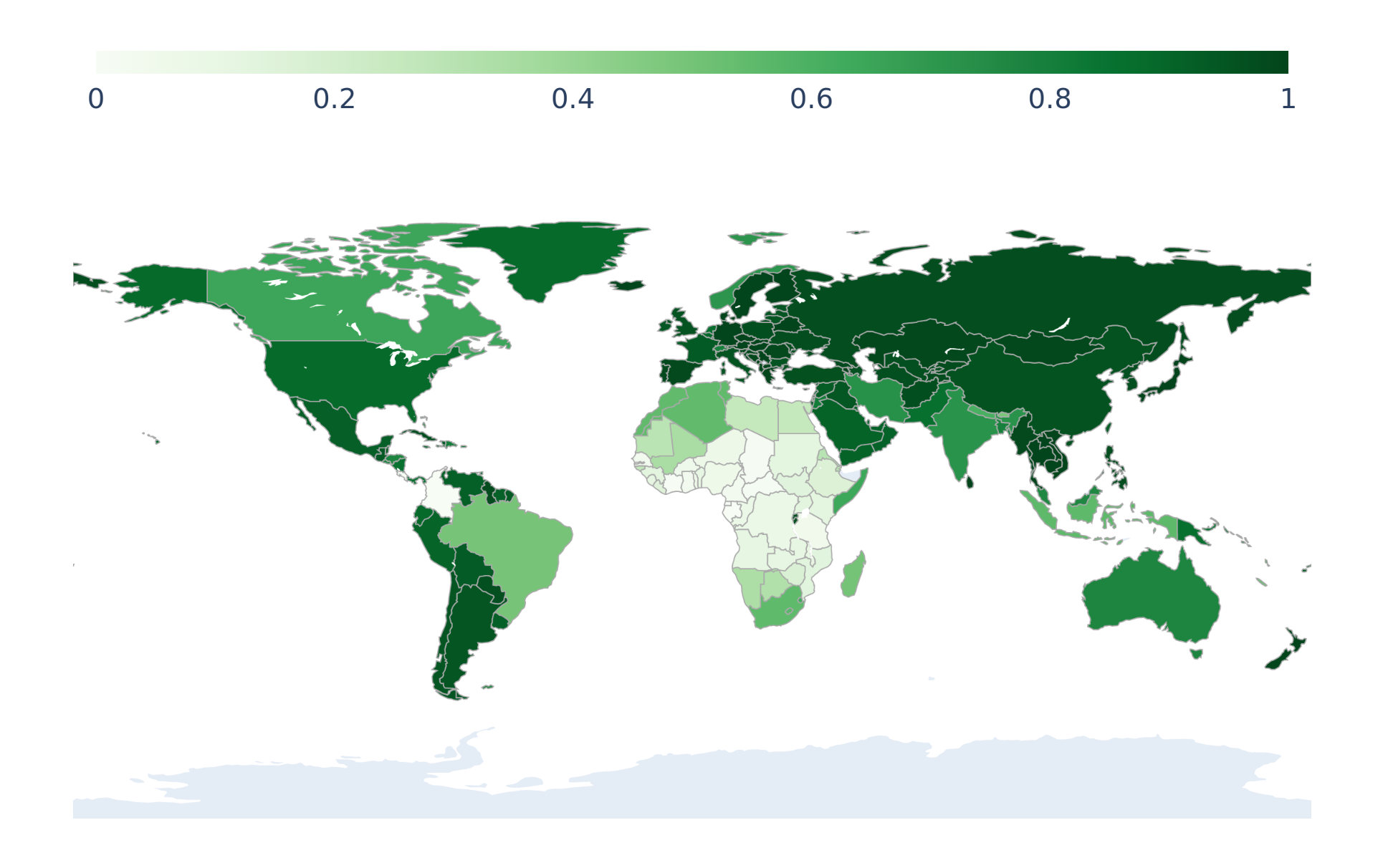}
  \caption{A larger choropleth map where the intensity indicates the averaged F\textsubscript{1} score of languages spoken in each region. It can been that the support of languages has geographical discrepancy with African languages being less supported. The figure is drawn based on the results of one of our experimental setting: (Language Name Prompt [Alias-Dialect-accepting], GPT-4, hard, 0-shot)}
  \label{fig:language_world_map_large}
 \end{centering}
\end{figure*}

\begin{table*}[h!]
    \begin{centering}
    \begin{tabular}{c}
    \begin{minipage}{1\textwidth}
      \includegraphics[width=0.99\linewidth]{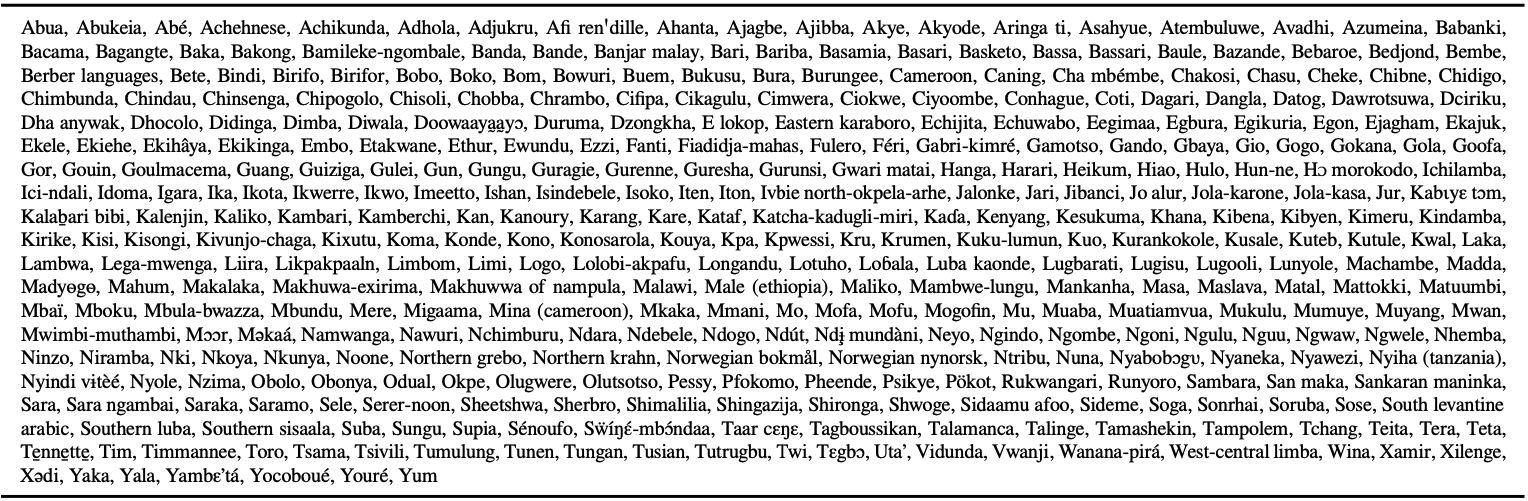}
    \end{minipage}
    \end{tabular}
    \caption{Languages with zero F1 scores in (Language name prompt[alias-dialect-accepting], GPT-4, hard, 0-shot) setting. The languages are ordered in alphabetical order. }
    \label{tab:lang_names_zero_f1_LNP_dialect_accepting}
    \end{centering}
\end{table*}

\begin{table}[h]
\scriptsize
\centering
\begin{tabular}{cccp{7.5cm}}
\toprule
\textbf{Artifect} & \textbf{Version} & \textbf{License} & \textbf{URL}  \\
\midrule
AfroLID & 2.1  & Apache 2.0 & \url{https://github.com/UBC-NLP/afrolid} \\
CLD2 & 0.41 & Apache 2.0 & \url{https://github.com/aboSamoor/pycld2} \\
CLD3 & 0.22 & Apache 2.0 & \url{https://github.com/bsolomon1124/pycld3} \\
FastText & 0.9.2 & MIT & \url{https://github.com/facebookresearch/fastText} \\
Franc & 6.1.0 & MIT & \url{https://github.com/wooorm/franc} \\
LangDetect & 1.0.9 & MIT & \url{https://github.com/fedelopez77/langdetect} \\
Langid.py &1.16 & Copyright 2011 Marco Lui & \url{https://github.com/saffsd/langid.py} \\

\bottomrule
\end{tabular}

\caption{Artifacts information. Our use of artifacts is consistent with their intended use, based on each their licenses.}  \label{tab:artifacts_info}
\end{table}
\normalsize
\clearpage
\section{Prompt Template and Prompt Examples}\label{sec:prompt_template_and_prompt_examples}
This appendix contains prompt templates and examples.  Table~\ref{tab:LNP_LCP_template} presents templates for Language Name Prompt (LNP) and Language Code Prompt (LCP). Tables~\ref{tab:LNP_examples_easy}, ~\ref{tab:LNP_examples_medium}, and~\ref{tab:LNP_examples_hard} showcase actual prompt examples for LNP at easy, medium, and hard diﬀiculty levels for LNP, respectively. Similarly, Tables~\ref{tab:LCP_examples_easy}, ~\ref{tab:LCP_examples_medium}, and ~\ref{tab:LCP_examples_hard}  are for LCP.

\begin{table*}[h!]
    \begin{centering}
    \begin{tabular}{c}
    \begin{minipage}{1\textwidth}
      \includegraphics[width=0.99\linewidth]{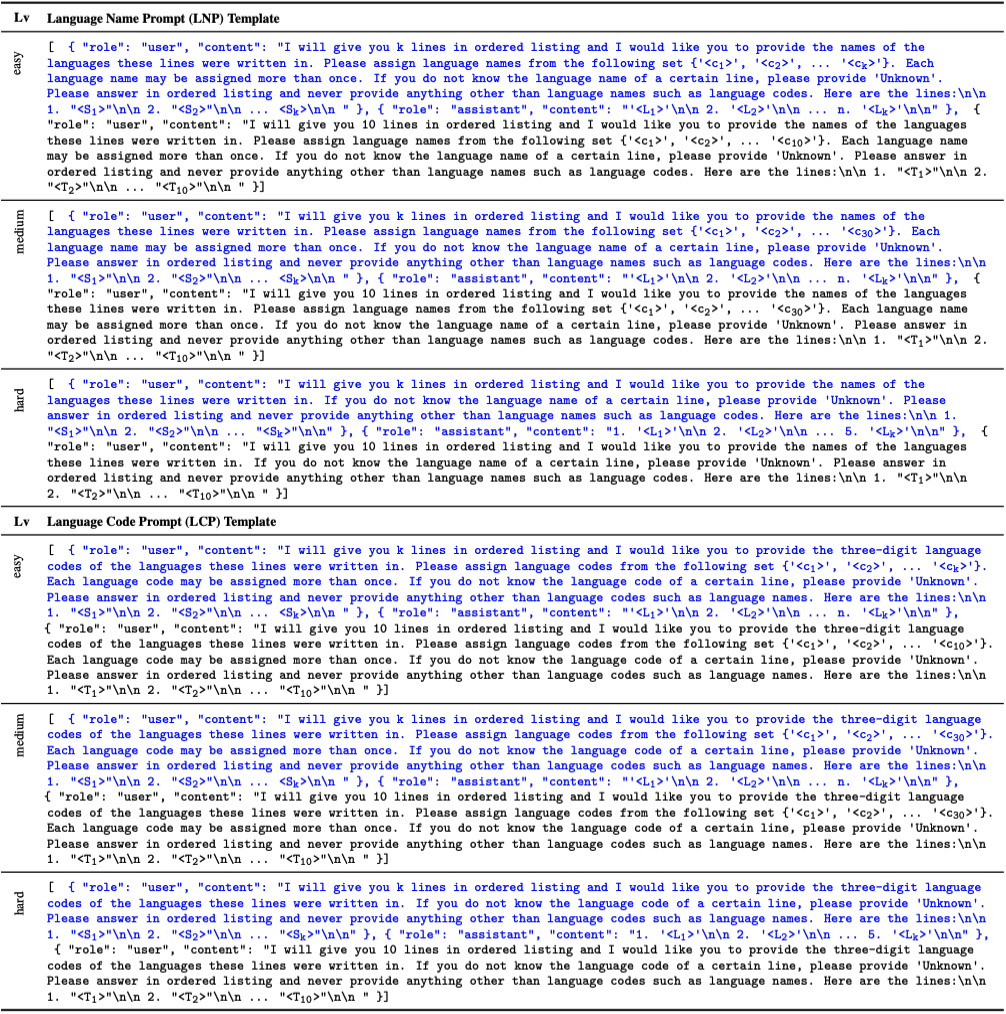}
    \end{minipage}
    \end{tabular}
    \caption{Language name prompt (LNP) and language code prompt (LCP) templates of k-shot for three difficulty levels (Lv). <S\textsubscript{i}>, <c\textsubscript{i}>, <L\textsubscript{i}>, <T\textsubscript{i}> are placeholders for i\textsuperscript{th} shot, i\textsuperscript{th} class label, i\textsuperscript{th} ground truth label for the i\textsuperscript{th} shot, i\textsuperscript{th} test example, respectively. <c\textsubscript{i}> and  <L\textsubscript{i}> are both language names for LNP templates and both language codes for LCP templates. The text in blue is for demonstration learning which shows ChatGPT the format and the content of our question and the expected response. The blue text will not be present for zero-shot setting. The system message is "You are a system which performs language identification." for all settings.}
    \label{tab:LNP_LCP_template}
    \end{centering}
\end{table*}

\begin{table*}[h!]
    \begin{centering}
    \begin{tabular}{c}
    \begin{minipage}{1\textwidth}
      \includegraphics[width=0.99\linewidth]{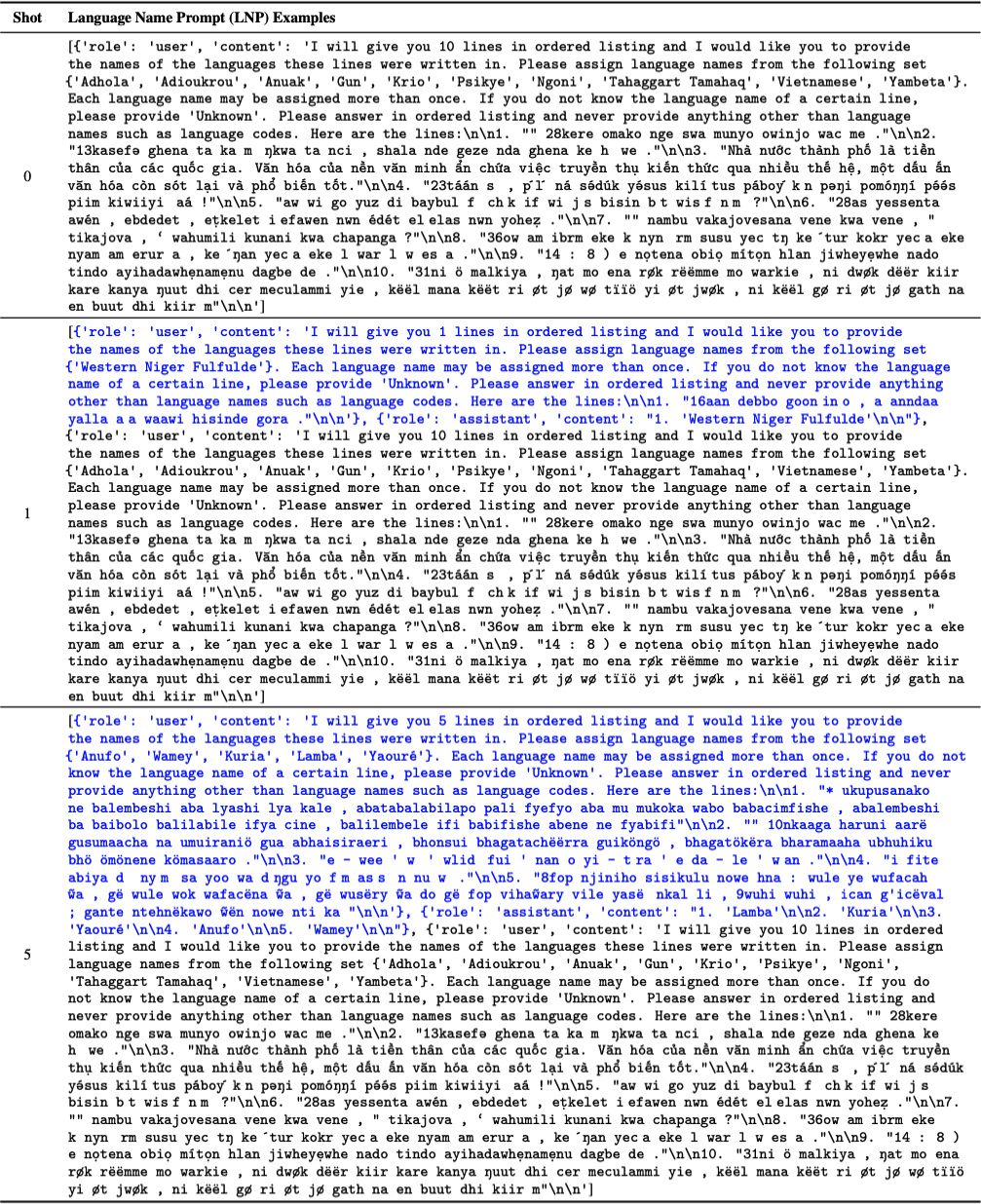}
    \end{minipage}
    \end{tabular}
    \caption{Request Examples of language name prompt (LNP) under three different numbers of shot at easy level where a label set of size equal to the number of unique language names of the test examples (i.e. $\sim10$) is provided to ChatGPT. The text in blue is for demonstration learning which shows ChatGPT the format and the content of our question and the expected response. We try to avoid harmful content by using Google Translate to translate its supported languages to English to inspect. However, not all languages included in a batch are supported by Google Translate. Therefore, the example may unintentionally include harmful content.}
    \label{tab:LNP_examples_easy}
    \end{centering}
\end{table*}

\begin{table*}[h!]
    \begin{centering}
    \begin{tabular}{c}
    \begin{minipage}{1\textwidth}
      \includegraphics[width=0.99\linewidth]{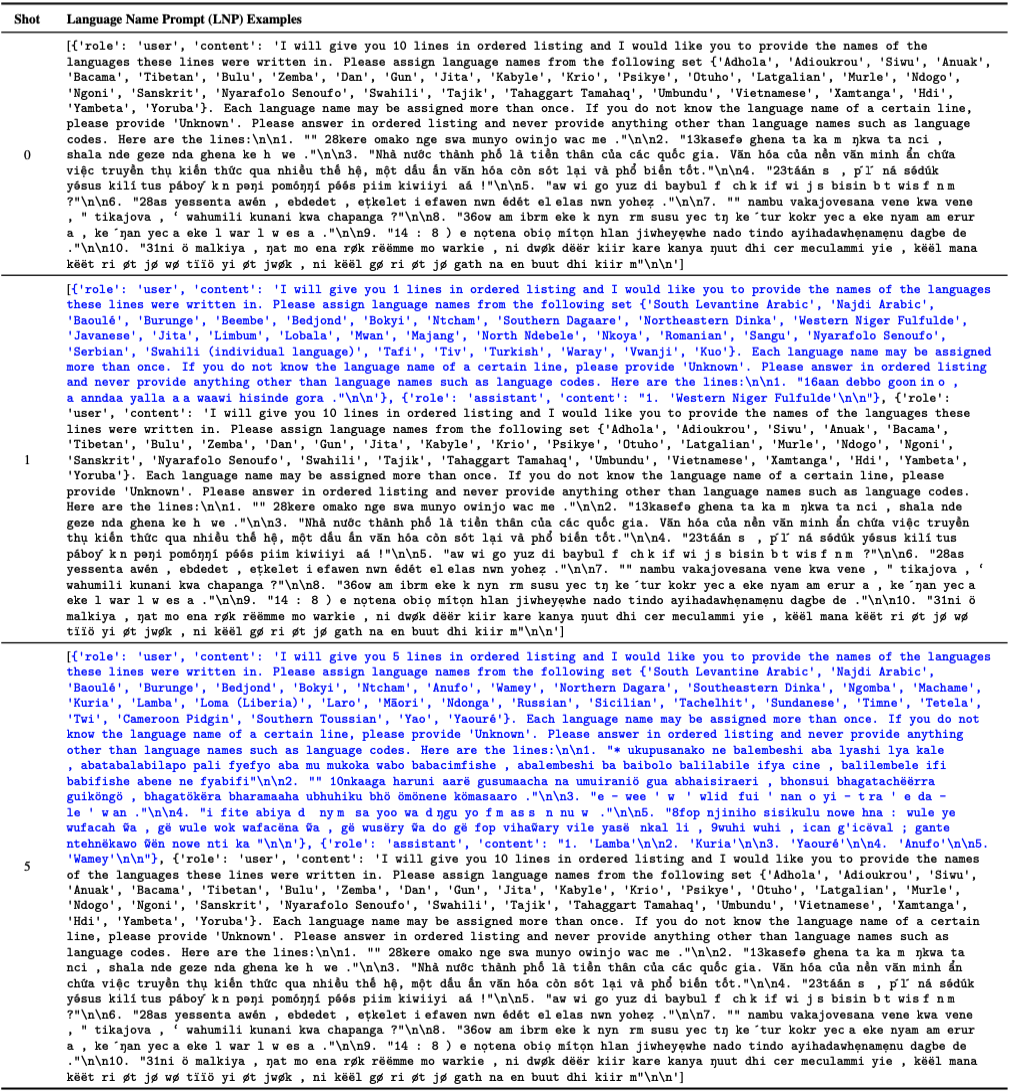}
    \end{minipage}
    \end{tabular}
    \caption{Request Examples of language name prompt (LNP) under three different numbers of shot at medium level where a label set of size $30$ is provided to ChatGPT. The text in blue is for demonstration learning which shows ChatGPT the format and the content of our question and the expected response. We try to avoid harmful content by using Google Translate to translate its supported languages to English to inspect. However, not all languages included in a batch are supported by Google Translate. Therefore, the example may unintentionally include harmful content.}
    \label{tab:LNP_examples_medium}
    \end{centering}
\end{table*}

\begin{table*}[h!]
    \begin{centering}
    \begin{tabular}{c}
    \begin{minipage}{1\textwidth}
      \includegraphics[width=0.99\linewidth]{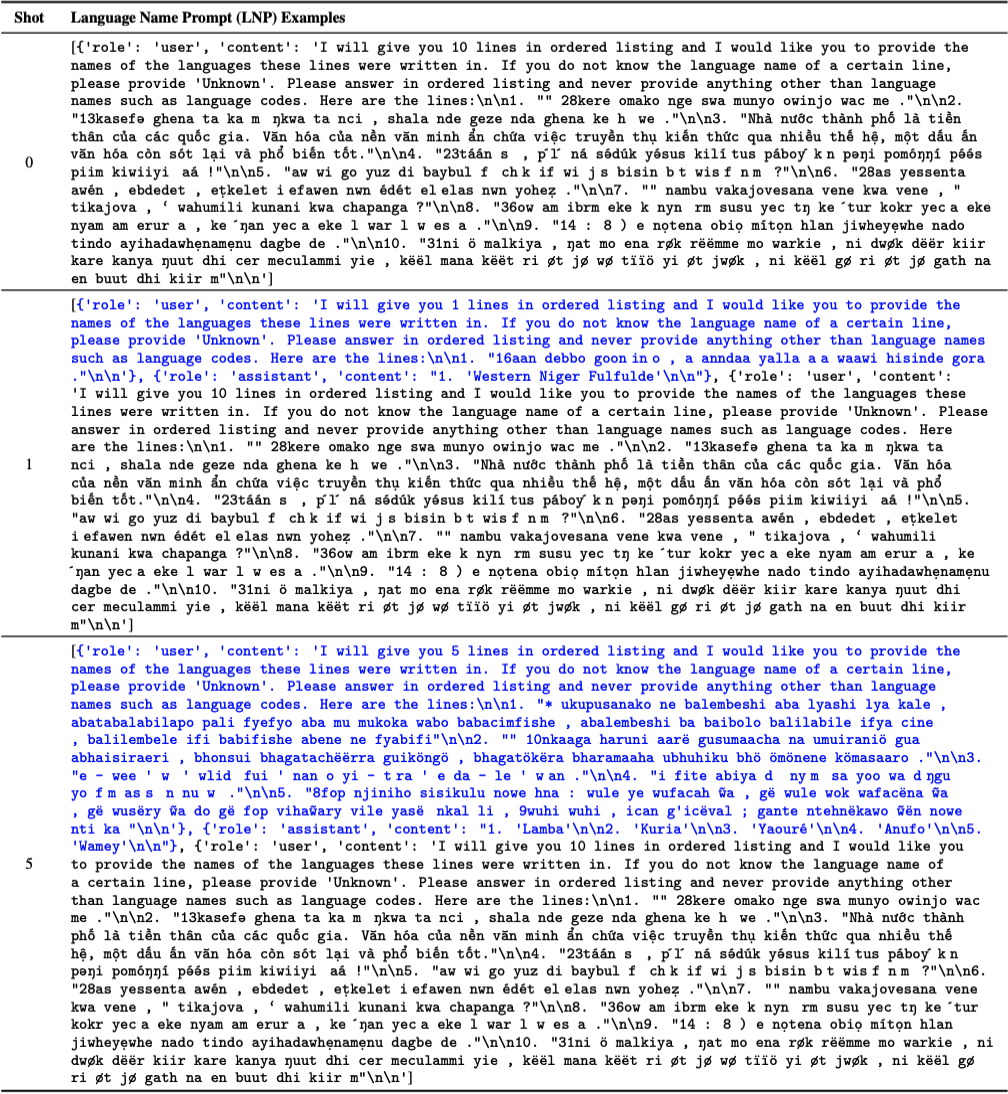}
    \end{minipage}
    \end{tabular}
    \caption{Request Examples of language name prompt (LNP) under three different numbers of shot at hard level where no label set is provided to ChatGPT. The text in blue is for demonstration learning which shows ChatGPT the format and the content of our question and the expected response. We try to avoid harmful content by using Google Translate to translate its supported languages to English to inspect. However, not all languages included in a batch are supported by Google Translate. Therefore, the example may unintentionally include harmful content.}
    \label{tab:LNP_examples_hard}
    \end{centering}
\end{table*}

\begin{table*}[h!]
    \begin{centering}
    \begin{tabular}{c}
    \begin{minipage}{1\textwidth}
      \includegraphics[width=0.99\linewidth]{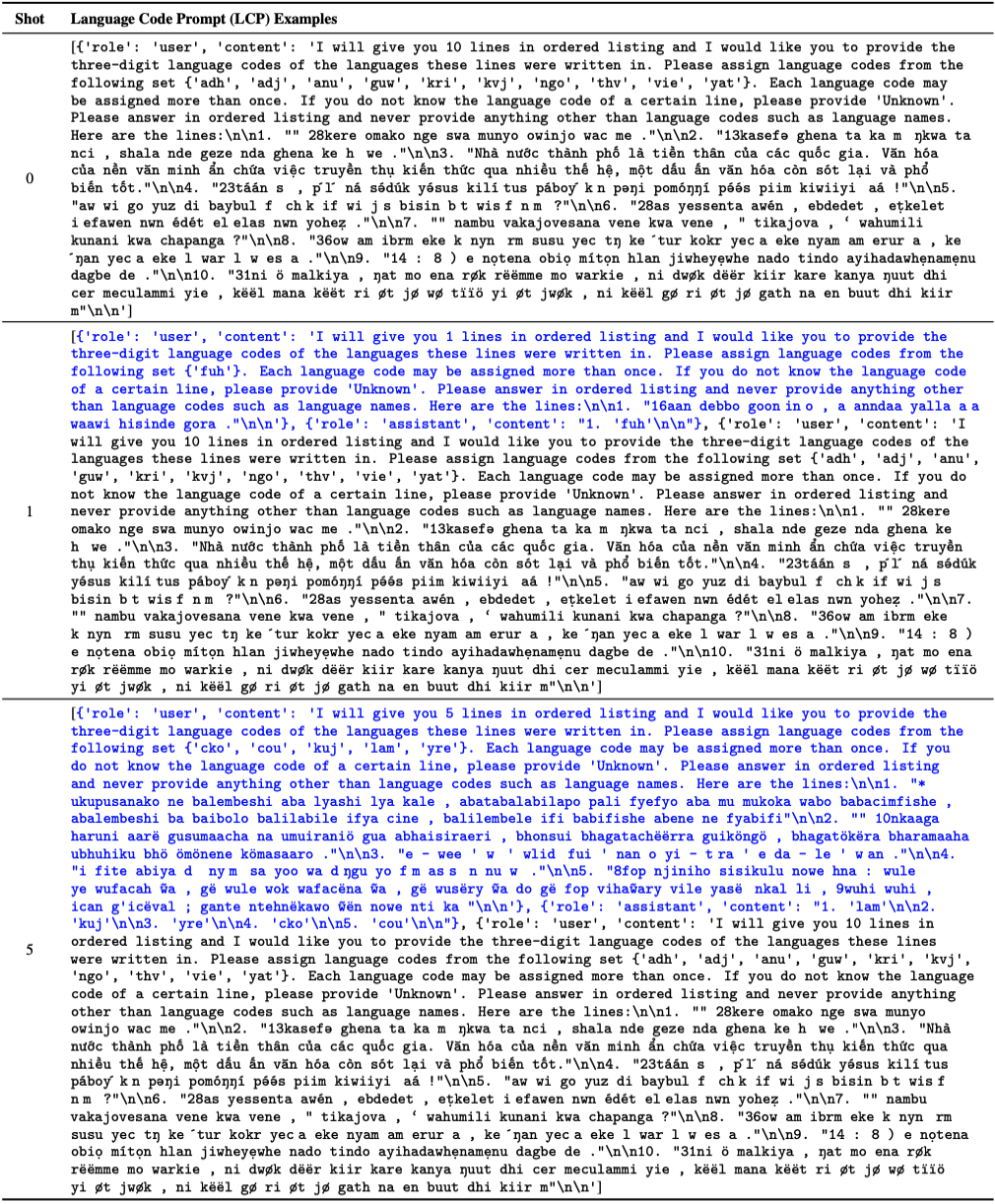}
    \end{minipage}
    \end{tabular}
    \caption{Request Examples of language code prompt (LCP) under three different numbers of shot at easy level where a label set of size equal to the number of unique language names of the test examples (i.e. $\sim10$) is provided to ChatGPT. The text in blue is for demonstration learning which shows ChatGPT the format and the content of our question and the expected response. We try to avoid harmful content by using Google Translate to translate its supported languages to English to inspect. However, not all languages included in a batch are supported by Google Translate. Therefore, the example may unintentionally include harmful content.}
    \label{tab:LCP_examples_easy}
    \end{centering}
\end{table*}

\begin{table*}[h!]
    \begin{centering}
    \begin{tabular}{c}
    \begin{minipage}{1\textwidth}
      \includegraphics[width=0.99\linewidth]{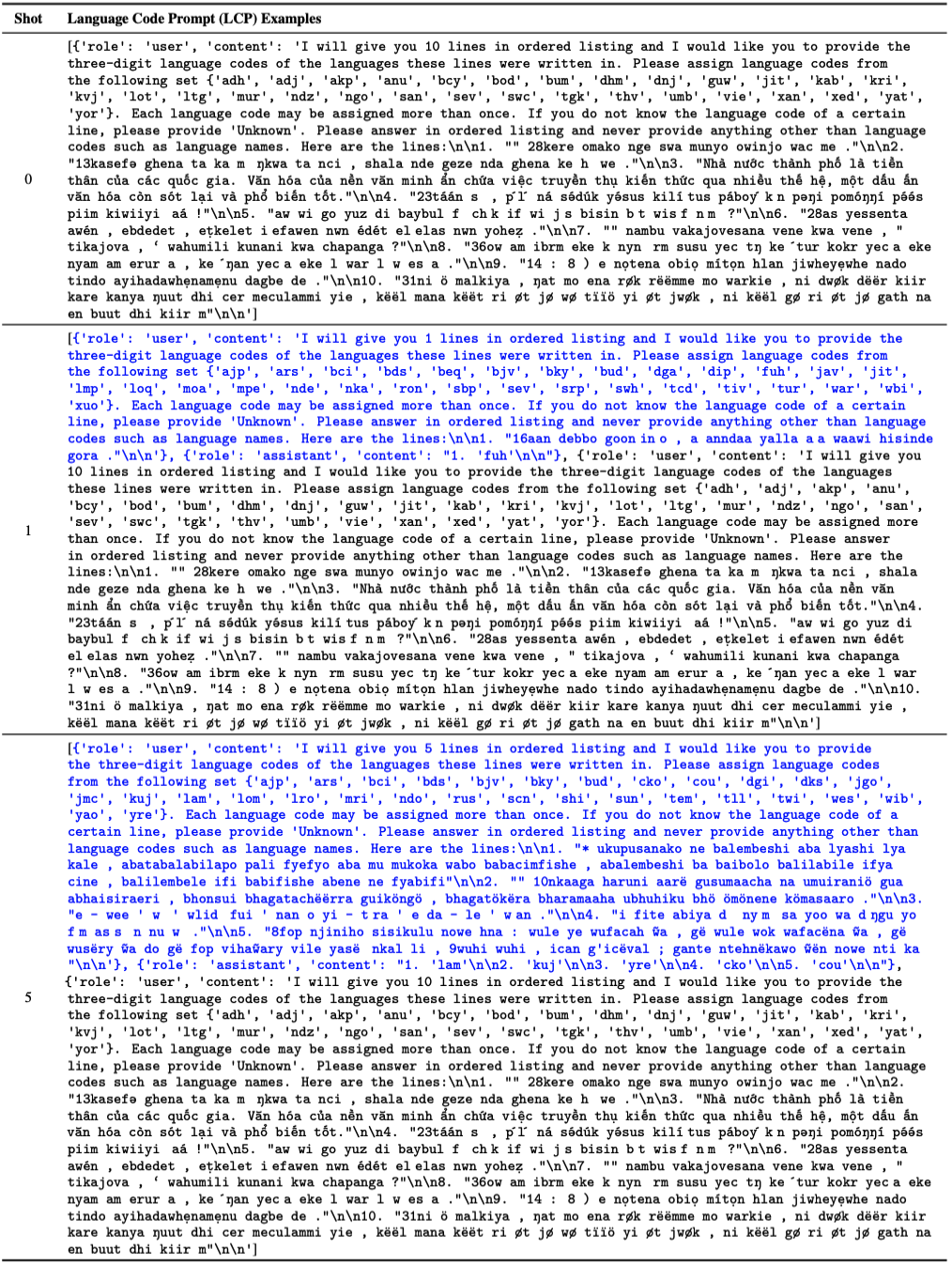}
    \end{minipage}
    \end{tabular}
    \caption{Request Examples of language code prompt (LCP) under three different numbers of shot at medium level where a label set of size $30$ is provided to ChatGPT. The text in blue is for demonstration learning which shows ChatGPT the format and the content of our question and the expected response. We try to avoid harmful content by using Google Translate to translate its supported languages to English to inspect. However, not all languages included in a batch are supported by Google Translate. Therefore, the example may unintentionally include harmful content.}
    \label{tab:LCP_examples_medium}
    \end{centering}
\end{table*}

\begin{table*}[h!]
    \begin{centering}
    \begin{tabular}{c}
    \begin{minipage}{1\textwidth}
      \includegraphics[width=0.99\linewidth]{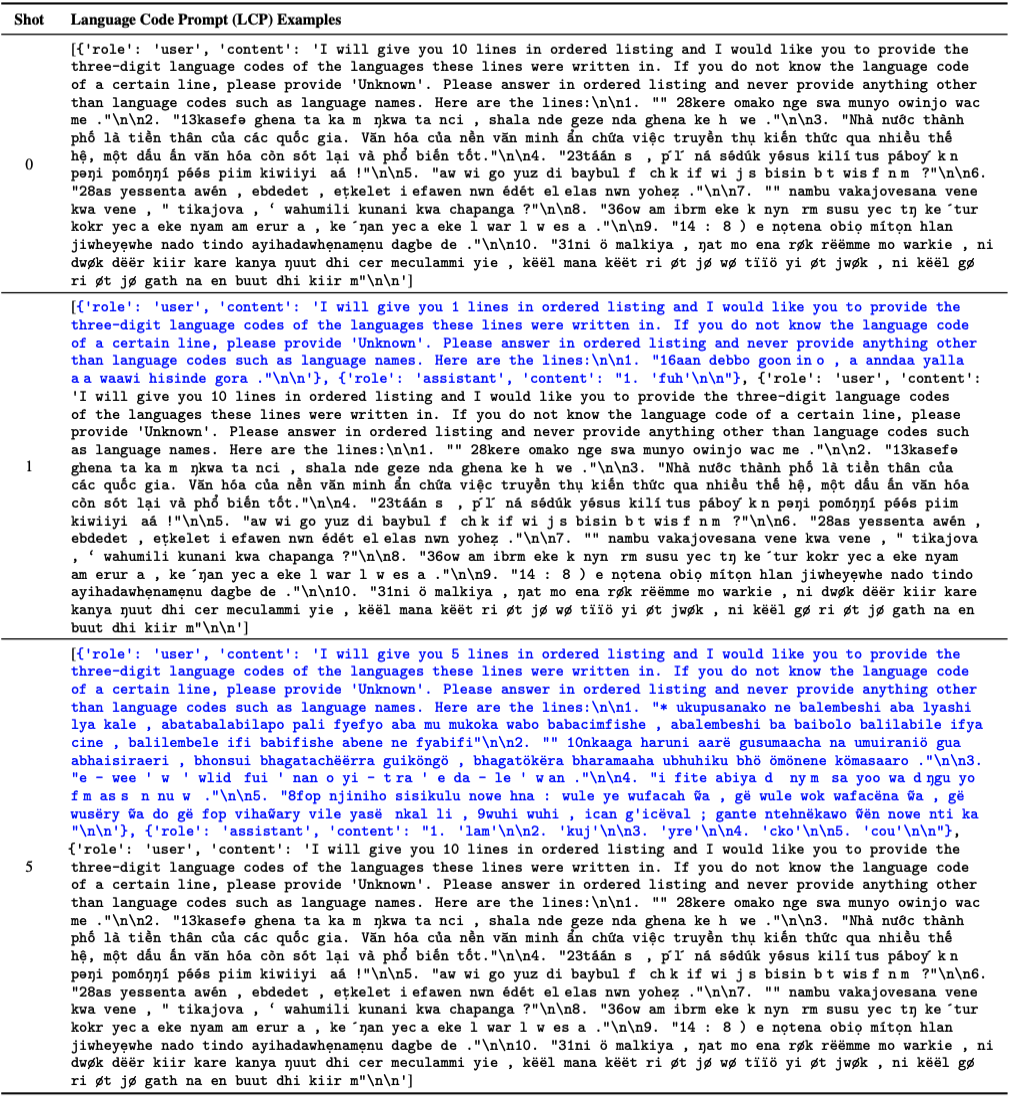}
    \end{minipage}
    \end{tabular}
    \caption{Request Examples of language code prompt (LCP) under three different numbers of shot at hard level where no label set is provided to ChatGPT. The text in blue is for demonstration learning which shows ChatGPT the format and the content of our question and the expected response. We try to avoid harmful content by using Google Translate to translate its supported languages to English to inspect. However, not all languages included in a batch are supported by Google Translate. Therefore, the example may unintentionally include harmful content.}
    \label{tab:LCP_examples_hard}
    \end{centering}
\end{table*}
\clearpage
\section{Alias-Dialect-Accepting Evaluation}\label{sec:ada_evaluation}

\begin{figure*}[htp!]
\begin{centering}
\includegraphics[scale=0.17]{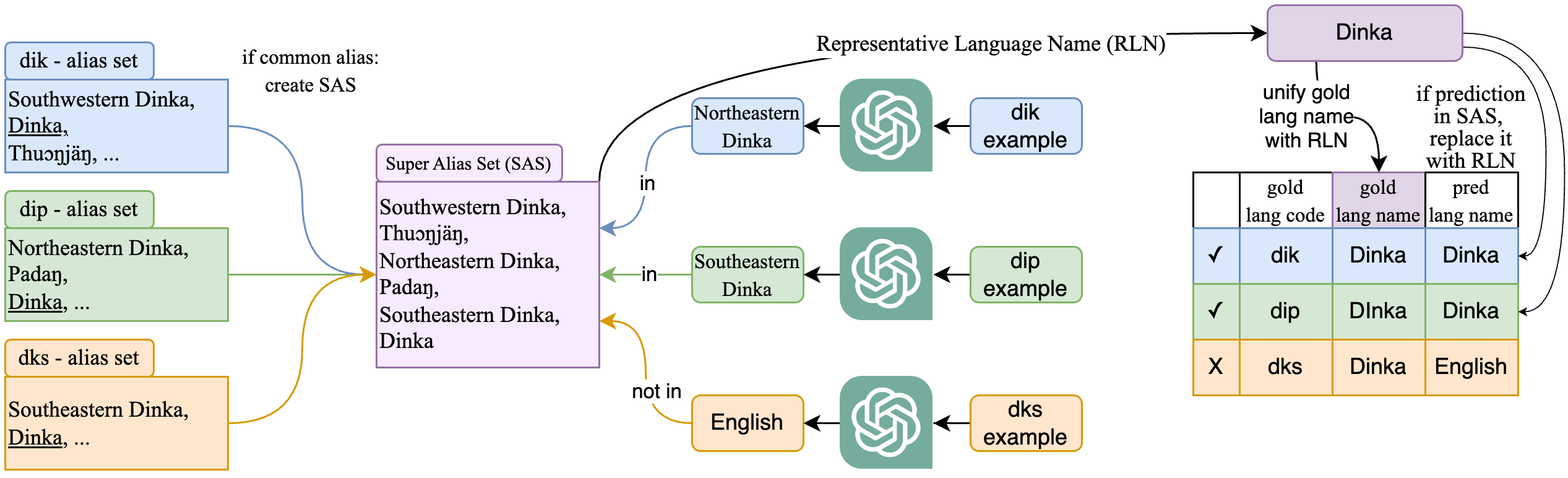}
  \caption{An overview of alias-dialect-accepting (ADA) evaluation for language name prompt (LNP). }
  \label{fig:ada_overview}
 \end{centering}
\end{figure*}

When prompting ChatGPT to predict language names, exact-match evaluation may not be the best approach to assess its language identification ability as dicussed in Section~\ref{sec:evaluation}. We propose alias-dialect-accepting evaluation, which counts a prediction to be a hit if it is an alias or a dialect that belongs to the same language group as ground truth label, to provide a fuzzy matching strategy. We introduce the two main components of this evaluation methods in two sections: accepting aliases in \ref{sec:accepting_aliases} and accepting dialects in \ref{sec:accepting_dialects}. An overview can be seen at Figure~\ref{fig:ada_overview}.

\subsection{Accepting Aliases}\label{sec:accepting_aliases}
A language can have more than one name, e.g. Español and Spanish. In fact, many languages have more than one name (636 out of 670 languages in Babel-670). The set of name(s) that belongs to a language code is referred to as the \textbf{alias set}. For example, for language code `spa', its alias set is $\{\text{Spanish, Español, Castellano, Castilian}\}$. To build an alias set for each language, we use the language code as unique identifier to consult \textit{Ethnologue} and the Python package \verb|langcodes|\footnote{https://github.com/rspeer/langcodes}. An alias set is a union of the set from Ethnologue and that from \verb|langcodes|. For Ethnologue, we take the values in three fields in each language page: \textit{Language Name}, \textit{Alternate Names} and \textit{Autonym}~\footnote{Following the definition of Ethnologue, an autonym is `the name of the language in the language itself'. For example, Español and Castellano are autonyms of Spanish}. For \verb|langcodes|, language name is retrieved by \verb|Language.get(<code>).language_name()|. For exact-match evaluation, if ChatGPT predicts `Español' when the label is `Spanish', it is counted a miss despite that they refer to the same language entity. With the design of accepting aliases, predicting `Español' will be counted as a hit for a Spanish example since `Español' is a member of the alias set of the language code `spa'. 

Besides names such as Spanish and Español which both refer to the same language entity, some alias sets include names that are referring to a group of languages. For example, the alias set of dik (Southwestern Dinka) includes `Dinka', which refers to a group of languages. This phenomenon often occurs when the language code belongs to a macrolanguage, e.g. dik (Southwestern Dinka) belongs to a macrolanguage Dinka (din). Besides, dip (Northeastern Dinka) and dks (Southeastern Dinka) both have `Dinka' in their alias sets and belong to macrolanguage Dinka.

Besides theoretical motivations as mentioned above, we have empirical motivations to accept aliases. We observe that ChatGPT tends to be conservative for languages that belong to a language group by predicting only its group name without giving dialectal information. For example, out of all test examples of dik, $87\%$ of them are predicted as `Dinka', instead of the more detailed `Southwestern Dinka' that includes dialectal information. Similar phenomenon occurs for a wide range of languages as can be seen in Table~\ref{tab:granular_alias_accepting_vs_exact_match}. We observe that some of these languages are clustered into groups, e.g. Dinka, Arabic, Kurdish, and Azerbaijani. These language groups include members that are linguistically related to each other. The grouping effect inspires us to propose to accept dialects which is covered in Section~\ref{sec:accepting_dialects}. 

\begin{table*}[htp!]
\tiny
\centering
\begin{tabular}{cccr}
\toprule
\textbf{Lang Code} & \textbf{Most Predicted Name} & \textbf{Gold Label} & \textbf{Rate}\\
\midrule
dik & Dinka & Southwestern Dinka & 87\%\\
dip & Dinka & Northeastern Dinka & 73\%\\
dks & Dinka & Southeastern Dinka & 53\%\\
acm & Arabic & Mesopotamian Arabic & 100\%\\
ars & Arabic & Najdi Arabic & 100\% \\
apc & Arabic & North Levantine Arabic & 100\%\\
acq & Arabic & Ta'izzi-Adeni Arabic & 93\%\\
aeb & Arabic & Tunisian Arabic & 87\%\\
arz & Arabic & Egyptian Arabic & 73\%\\
ary & Arabic & Moroccan Arabic & 53\%\\
kmr & Kurdish & Northern Kurdish & 100\%\\
ckb & Kurdish & Central Kurdish & 93\%\\
azb & Azerbaijani & South Azerbaijani & 100\%\\
azj & Azerbaijani & North Azerbaijani & 100\%\\
pes & Persian & Iranian Persian & 100\%\\
ydd & Yiddish & Eastern Yiddish & 100\%\\
khk & Mongolian & Halh Mongolian & 100\%\\
ayr & Aymara & Central Aymara & 87\%\\
als & Albanian & Tosk Albanian & 100\%\\
gaz & Oromo & West Central Oromo & 100\%\\
plt & Malagasy & Plateau Malagasy & 100\%\\
uzn & Uzbek & Northern Uzbek & 100\%\\
yue & Chinese & Cantonese & 100\%\\
pbt & Pashto & Southern Pashto & 93\%\\
quy & Quechua & Ayacucho Quechua & 93\%\\
\bottomrule
\end{tabular}

\caption{Languages with more than half of their predictions ($\ge8$ datapoints as each language has $15$ test examples) considered as misses in exact-match evaluation, but considered as hits when aliases are accepted (i.e. when prediction is a member of the alias set). \textit{Most Predicted Name} is the language name that is predicted by ChatGPT most frequently. \textit{Rate} refers to the ratio of the number of these predictions, out of 15 test examples. This analysis is based on (LNP, GPT-4, hard, 0-shot) setting. }
\label{tab:granular_alias_accepting_vs_exact_match}
\end{table*}

\subsection{Accepting Dialects}\label{sec:accepting_dialects}
Following the observation of grouping effect when accepting alias as discussed in section~\ref{sec:accepting_aliases}, we propose to accept dialects and count a prediction as a hit if the ground truth label and the prediction are dialects of a common language. We take Southwestern Dinka as an example and illustrate how its example is counted as a hit when accepting dialect in  Figure~\ref{fig:ada_overview}. Implementation Details are shown in Algorithm~\ref{alg:dialect_accepting}.

First, we assume that whenever a common language name occurs in the alias sets of two languages, they are linguistically related in a certain degree. They will form a language group and their alias sets will be merged. For one language $x$, after iterating through each pair of $x$ and the other languages, a \textbf{super alias set (SAS)} of $x$ is formed by merging all the alias set(s) having at least one common language name with the alias set of $x$. After the merges, out of $670$ languages, $595$ language groups are formed with $48$ of them having more than one languages in a group, and $547$ of them having one single language in the group. Second, we select a \textbf{representative language name (RLN)} to represent each language group by picking the most frequently occurred name in the SAS (more details covered in Phase 2 in Algorithm~\ref{alg:dialect_accepting}). This is reasonable because we observe that the most common name among alias sets of a language group is often the name of the macrolanguage. Third, we use RLN as the gold label for a language group and replace any predicted names with the RLN if they are in the SAS. Fourth, we compute F\textsubscript{1} score for each language group and a macro-averaged F\textsubscript{1} to present the overall system performance. The macro-averaged (by language group) F\textsubscript{1} scores for (LNP [alias-dialect-accepting], hard, 0-shot) setting of both GPT-3.5 and GPT-4 are included in Table~\ref{tab:results}.

It is noteworthy a difference in accepting alias and accepting dialect. Accepting aliases allows a test example of Southwestern Dinka to be predicted as `Dinka' because `Dinka' is in its alias set. For the same example, accepting dialects allows it to be predicted as `Northeasten Dinka' because they are both dialects of Dinka and are merged under the RLN Dinka.

\begin{algorithm*}
\caption{Alias-Dialect-Accepting Evaluation Algorithm}\label{alg:dialect_accepting}
\label{alg:proximity_accepting}
\begin{algorithmic}[1]

\Require
    \Statex $AS_1, AS_2, \dots AS_n \textrm{ where $AS_i$ is the alias set, in multiset structure, of ith language (details in~\ref{sec:accepting_aliases})}$
    \Statex $Golds_1, Golds_2, \dots Golds_n \textrm{ where $Golds_i$ is a m-sized list of gold names of ith language}$
    \Statex $Preds_1, Preds_2, \dots Preds_n \textrm{ where $Preds_i$ is a m-sized list of predicted names of ith language}$

\phase{Create super alias set (SAS) and forming language groups}

\For{\texttt{i = 1 to n}}
    \State $SAS_i = AS_i$ \Comment{$SAS_i$ is also a multiset}
    \For{\texttt{j = 1 to n}}
        \If{\texttt{i==j}} 
            \State continue 
        \EndIf
        \If{$AS_i \cap AS_j \neq \emptyset$} \Comment{if at least one common language name exists}
            \State $SAS_i = SAS_i \uplus AS_j$ \Comment{add new aliases and multiplicity into SAS with multiset union}
        \EndIf
    \EndFor
\EndFor
\phase{Select representative language name (RLN) for each language group}
\State $Candidates_i = \underset{x}{\arg\max}(m_{SAS_{i}}(x))$ \Comment{Get name(s) with highest multiplicity as candidates of RLN}
\If{$|Candidates_i| == 1$} \Comment{if there is one single candidate, it is assigned as RLN}
    \State $RLN_i = Candidates_i[0]$
\Else \Comment{if more than one candidate, pick the most frequently predicted name that is in SAS}
    \State $CPN_i = \{name \hspace{0.1cm}| \hspace{0.1cm} name \in SAS_i \land name \in Preds_i\}$ \Comment{CPN: Correctly predicted name(s)}
    \If{$|CPN_i| == 0$} \Comment{if none of the predictions is in SAS}
        \State $RLN_i \sim Uniform(Candidates_i)$ \Comment{Uniformly drawing one name out of $Candidates_i$} %
    \Else
        \State $MFCPN_i =  \underset{x}{\arg\max}(m_{CPN_{i}}(x))$ \Comment{MFCPN: Most frequent correctly predicted name(s)}
        \If{$|MFCPN_i| == 1$}
            \State $RLN_i = MFCPN_i[0]$
        \Else
            \State $RLN_i \sim Uniform(MFCPN_i)$ \Comment{Uniformly drawing one name out of $MFCPN_i$}
        \EndIf
    \EndIf
    
\EndIf

\phase{Replace gold and predicted name (if in SAS) with RLN for each datapoint}
\For{\texttt{i = 1 to n}}
    \For{\texttt{j = 1 to m}}
        \State $Golds_{i,j} = RLN_i$
        \If{$Preds_{i,j} \in SAS_i$}
            \State $Preds_{i,j} = RLN_i$
        \EndIf
    \EndFor
\EndFor

\phase{Evaluation of $F_1$ for each language group K and macro-averaged $F_1$ for the system}
\State $F_{1_K} = F_1(Golds_{k \in K}, Preds_{k \in K})$ \Comment{$k \in K$ represents member language's lists concatenated}

\State $F_{1_{sys}} = mean(\{F_{1_K}| 1 \leq K \leq N\})$ \Comment{N is the total number of language groups}

\end{algorithmic}
\end{algorithm*}
\clearpage
\section{Languages in Babel-670}\label{sec:langs_in_babel670}

\begin{table*}[h!]
\small
\centering
\resizebox{\textwidth}{!}{%
\begin{tabular}{ll ll ll ll }
\toprule
\textbf{ISO-3} & \textbf{Language}                & \textbf{ISO-3} & \textbf{Language}             & \textbf{ISO-3} & \textbf{Language}                       & \textbf{ISO-3} & \textbf{Language}         \\
\midrule
aar  & Afar / Qafar& bky   & Bokyi & ego &   Eggon  & heb  &  Hebrew \\
aba   & Abe / Abbey & bmo & Bambalang  &eka&   Ekajuk & heh&Hehe  \\
abn   & Abua & bmv & Bum &eko& Koti & her&  Herero \\
acd   & Gikyode & bod   & Standard Tibetan &ell  &  Greek &hgm & Haillom   \\
ace & Acehnese  & bom & Berom  & eng & English & hin & Hindi\\
ach & Acholi & bos & Bosnian & epo  & Esperanto& hna & Mina \\
acm & Mesopotamian Arabic & bov & Tuwuli & est & Estonian& hne & Chhattisgarhi\\
acq & Ta’izzi-Adeni Arabic & box & Bwamu / Buamu & eto & Eton & hrv & Croatian\\
ada   & Dangme  & bqc   & Boko  & etu &   Ejagham & hun&Hungarian \\
adh   & Jopadhola / Adhola & bqj   & Bandial &  etx  &   Iten / Eten & hye & Armenian\\
adj   & Adjukru  / Adioukrou & bsc & Oniyan &eus & Basque& ibb& Ibibio     \\
aeb & Tunisian Arabic & bsp & Bagag Sitemu & ewe  &  Ewe  & ibo & Igbo\\
afr   & Afrikaans  & bss   & Akose   &ewo   & Ewondo & idu& Idoma\\
agq   & Aghem & bst   & Basketo &fak   & Fang & igb  & Ebira\\
aha   & Ahanta   & bud   & Ntcham   & fao & Faroese & ige& Igede \\
ajg   & Aja  & bug   & Buginese   &fat   & Fante  & igl& Igala        \\
ajp & South Levantine Arabic  & bul   & Bulgarian & ffm   & Fulfulde, Maasina& ijn &Kalabari\\
akp   & Siwu   & bum   & Bulu & fia &     Nobiin  & ikk &   Ika     \\
als & Tosk Albanian  & bun   & Sherbro & fij   &   Fijian & ikw & Ikwere\\
alz   & Alur  &buy   & Bullom So  & fin&Finnish & ilo & Ilocano    \\
amh   & Amharic  & bwr   & Bura Pabir   &  fip   & Fipa  &ind  &  Indonesian \\
ann   & Obolo  &  bwu  &  Buli  &flr   & Fuliiru  & iqw& Ikwo\\
anu   & Anyuak / Anuak  & bxk   &  Bukusu  & fon   & Fon  & iso& Isoko \\
anv   & Denya   &  byf  &   Bete  & fra  & French & isl&  Icelandic\\
apc & North Levantine Arabic & byv   & Medumba & fub   & Fulfulde, Adamawa& ita & Italian\\
arb & Modern Standard Arabic & bza   & Bandi & fue   & Fulfulde, Borgu& iyx & Yaka\\
ars & Najdi Arabic & bzd & Bribri & fuf   & Pular & izr & Izere\\
ary & Moroccan Arabic  & bzw   & Basa & fuh   & Fulfulde, Western Niger & izz & Izii\\
arz & Egyptian Arabic & cat & Catalan & ful   & Fulah & jav & Javanese\\
asa & Asu  & cce &     Chopi  &fur  & Friulian & jgo&   Ngomba       \\
asg & Cishingini & ceb  &  Cebuano  &  fuv   & Fulfude Nigeria &jib & Jibu\\
asm & Assamese & ces & Czech & gaa   & Ga &jit &Jita\\
ast & Asturian & chw   & Chuabo & gax   & Oromo, Borana-Arsi-Guji &jmc & Machame\\
atg   & Ivbie North-Okpela-Arhe &  cjk  & Chokwe  &gaz   & Oromo, West Central &jpn & Japanese     \\
ati & Attie &  ckb  & Central Kurdish & gbo & Grebo, Northern & kab & Kabyle \\
avn & Avatime& cko  & Anufo & gbr & Gbagyi  &  kac& Jingpho  \\
avu   & Avokaya &  cme  & Cerma & gde   & Gude &kam & Kikamba    \\
awa & Awadhi  & cop   & Coptic & gid   & Gidar & kan & Kannada\\
ayr & Central Aymara & crh & Crimean Tatar & giz   & South Giziga & kas & Kashmiri\\
azb & South Azerbaijani & crs   & Seychelles &  gjn   & Gonja &kat & Georgian\\
azj & North Azerbaijani & csk   & Jola Kasa & gkn   & Gokana &kaz & Kazakh\\
azo & Awing & cwe  &  Kwere  &  gkp   & Kpelle, Guinea &kbn & Kare \\
bak & Bashkir & cym & Welsh & gla & Scottish Gaelic & kbo & Keliko\\
bam   & Bambara  & daa & Dangaleat  &gle  &Irish  & kbp &  Kabiye \\
ban & Balinese   & dag   & Dagbani & glg & Galician & kby &  Kanuri, Manga\\
bav   & Vengo& dan &  Danish & gmv   & Gamo &kcg& Tyap       \\
bba   & Baatonum  & dav & Dawida / Taita & gna & Kaansa &kck  &Kalanga \\
bbj   & Ghomala & deu & German  & gnd   & Zulgo-gemzek & kdc & Kutu\\
bbk   & Babanki & dga   & Dagaare & gng   & Ngangam & kde& Makonde \\
bcn   & Bali & dgd   & Dagaari Dioula &  gof   & Goofa& kdh& Tem\\
bcw   & Bana & dgi   & Dagara, Northern &  gog   & Gogo & kdi & Kumam\\
bcy   & Bacama & dhm   & Dhimba & gol & Gola & kdj& Ng’akarimojong\\
bdh   & Baka & dib & Dinka, South Central & gqr& Gor & kdl& Tsikimba \\
bds   & Burunge &  did   & Didinga & grn& Guarani &  kdn& Kunda\\
bel & Belarusian &  dig   & Chidigo & gso & Gbaya, Southwest& kea & Kabuverdianu\\
bem   & Bemba / Chibemba & dik   & Dinka, Southwestern & gud &  Dida, Yocoboue&  ken& Kenyang\\
ben & Bengali &   dip   & Dinka, Northeastern & guj & Gujarati & khk & Halh Mongolian\\
beq   & Beembe  &  diu   & Gciriku  & gur& Farefare&  khm& Khmer\\
ber   & Berber & dks   & Dinka, Southeastern &guw&  Gun &khy & Kele / Lokele\\
bex   & Jur Modo & dnj   & Dan & gux & Gourmanchema&kia &  Kim\\
bez   & Bena &   dow   & Doyayo &guz   & Ekegusii&kik &Gikuyu / Kikuyu    \\
bfa   & Bari &  dsh   & Daasanach & gvc& Wanano& kin& Kinyarwanda\\
bfd   & Bafut & dua   & Douala  & gvl & Gulay &kir & Kyrgyz\\
bfo   & Birifor, Malba &dug & Chiduruma & gwr&Gwere& kiz&Kisi  \\
bho & Bhojpuri & dwr   & Dawro &gya & Gbaya, Northwest & kkl & Kagulu\\
bib   & Bisa &dyi   & Sénoufo, Djimini & hag & Hanga &kkj & Kako  \\

\bottomrule
\end{tabular}%
}

\caption{List of languages in Babel-670 - Part I. }
\label{tab:lang_listI}
\end{table*}

\begin{table*}[]
\small
\centering
\resizebox{\textwidth}{!}{%
\begin{tabular}{ll ll ll ll }
\toprule
\textbf{ISO-3} & \textbf{Language}                & \textbf{ISO-3} & \textbf{Language}             & \textbf{ISO-3} & \textbf{Language}                       & \textbf{ISO-3} & \textbf{Language}         \\
\midrule
bim   & Bimoba &  dyu   & Jula&  har& Harari & kln&  Kalenjin  \\
bin   & Edo &dzo  &  Dzongkha &  hat&  Haitian Creole& klu& Klao \\
biv   & Birifor, Southern & ebr &  Ebrie& hau& Hausa& kma& Konni \\
bjn & Banjar & ebu & Kiembu / Embu & hay & Haya &kmb & Kimbudu\\
bjv   & Bedjond &efo & Efik &hbb &Nya Huba &kmr & Northern Kurdish   \\
kmy & Koma & lmp & Limbum &mfz & Mabaan & ndv & Ndut  \\
knf & Mankanya & lnl & Banda, South Central & mgc & Morokodo &  ndz & Ndogo \\
kng & Kongo  & log & Logo & mgh & Makhuwa-Meetto& ngb & Ngbandi, Northern \\
knk & Kuranko & lom & Loma & mgo & Meta' & ngc & Ngombe \\
kno & Kono & loq & Lobala &mgq & Malila & ngl & Lomwe \\
koo & Konzo& lot & Latuka &mgr & Mambwe-Lungu& ngn & Bassa  \\
koq & Kota& loz & Silozi &mgw & Matumbi & ngo & Ngoni\\
kor & Korean& lmo & Lombard &mif & Mofu-Gudur& ngp & Ngulu  \\
kqn & Kikaonde& lro & Laro &min & Minangkabau & nhr & Naro \\
kqp & Kimré & lsm & Saamya-Gwe / Saamia & mkd& Macedonian& nhu & Noone \\
kqs & Kisi & ltg & Latgalian &mkl & Mokole& nih & Nyiha \\
kqy & Koorete& lth & Thur / Acholi-Labwor & mlg & Malagasy& nim & Nilamba / kinilyamba \\
kri & Krio & lto & Tsotso & mlr & Vame & nin & Ninzo \\
krs & Gbaya & ltz & Luxembourgish & mlt & Maltese & niy & Ngiti  \\
krw & Krahn, Western& lua & Tshiluba & mmy & Migaama & nka & Nkoya / ShiNkoya \\
krx & Karon& luc & Aringa & mnf & Mundani & nko & Nkonya\\
ksb & Shambala / Kishambala& lue & Luvale &mnk & Mandinka &  nla & Ngombale \\
ksf & Bafia & lug & Luganda & mni & Meitei & nld & Dutch\\
ksp & Kabba & lun & Lunda & moa & Mwan &  nnb & Nande / Ndandi\\
ktj & Krumen, Plapo & luo & Dholuo/ Luo & mos & Moore &  nnh & Ngiemboon \\
ktu & Kikongo & lus & Mizo & moy & Shekkacho & nno & Norwegian Nynorsk \\
kua & Oshiwambo & lwg & Wanga / Saamia& moz & Mukulu & nnq & Ngindo  \\
kub & Kutep &lwo &Luwo & mpe & Majang & nob & Norwegian Bokmål\\
kuj & Kuria& lvs & Standard Latvian & mpg & Marba & npi & Nepali\\
kus & Kusaal & maf & Mafa & mqb & Mbuko & nse & Chinsenga\\
kvj & Psikye & mag & Magahi& mri & Maori &   nnw &  Nuni, Southern \\
kwn & Kwangali&mai & Maithili &  msc & Maninka, Sankaran & nso & Sepedi\\
kyf & Kouya & mal& Malayalam& mur & Murle & ntr & Delo\\
kyq & Kenga & mar & Marathi& muy & Muyang &  nuj & Nyole\\
kzr & Karang& mas& Maasai &  mwe & Mwera &   nus & Nuer \\
lai & Lambya& maw & Mampruli & mwm & Sar & nwb & Nyabwa\\
laj & Lango &  mbu & Mbula-Bwazza & mwn & Cinamwanga &   nxd & Ngando \\
lam & Lamba & mck & Mbunda& mws & Mwimbi-Muthambi & nya & Chichewa \\
lao & Lao & mcn & Masana / Massana & mya & Burmese & nyb & Nyangbo\\
lap & Laka &  mcp & Makaa &  myb & Mbay &  nyd & Olunyole / Nyore \\
lee & Lyélé& mcu & Mambila, Cameroon& myk & Sénoufo, Mamara & nyf & Giryama\\
lef & Lelemi &  mda & Mada& myx & Masaaba &  nyk & Nyaneka\\
lem &Nomaande &mdm & Mayogo & mzm & Mumuye & nym & Nyamwezi\\
lgg & Lugbara & mdy & Maale & mzw & Deg & nyn & Nyankore / Nyankole\\
lgm & Lega-mwenga &  men & Mende  & naq & Khoekhoe & nyo & Nyoro \\
lij & Ligurian& meq & Merey & naw & Nawuri & nyu & Nyungwe \\
lik & Lika & mer & Kimiiru & nba & Nyemba & nyy & Nyakyusa-Ngonde / Kyangonde\\
lim & Limburgish & mev & Maan / Mann& nbl & IsiNdebele &   nza & Mbembe, Tigon\\
lin & Lingala & mfe & Morisyen / Mauritian Creole &  nzi & Nzema & oci & Occitan\\
lip & Sekpele & mfg & Mogofin & ndc & Ndau  & odu & Odual \\
lit & Lithuanian & mfh & Matal & nde & IsiNdebele & ogo & Khana
\\
lla & Limba, West-Central & mfi & Wandala & ndh & Ndali&     oke & Okpe\\ 
lmd & Lumun & mfk & Mofu, North & ndj & Ndamba & okr & Kirike\\
lmo & Lombard & mfq & Moba & ndo & Ndonga & oku & Oku   \\
\bottomrule
\end{tabular}%
}
\caption{List of languages in Babel-670 - Part II }
\label{tab:lang_listII}
\end{table*}

\begin{table*}[]
\small
\centering
\resizebox{\textwidth}{!}{%
\begin{tabular}{ll ll ll ll }
\toprule
\textbf{ISO-3} & \textbf{Language}                & \textbf{ISO-3} & \textbf{Language}             & \textbf{ISO-3} & \textbf{Language}                       & \textbf{ISO-3} & \textbf{Language}         \\
\midrule
ncu                             & Chunburung                         & shk                             & Shilluk                            & teo                             & Teso                               & vmk                             & Makhuwa-Shirima                    \\
orm                             & Oromo                              & shn                             & Shan                               & tex                             & Tennet                             & vmw                             & Macua                              \\
ory                             & Odia                               & sid                             & Sidama                             & tgk                             & Tajik                              & vun                             & Kivunjo                            \\
ozm                             & Koonzime                           & sig                             & Paasaal                            & tgl                             & Tagalog                            & vut                             & Vute                               \\
pag                             & Pangasinan                         & sil                             & Sisaala, Tumulung                  & tgw                             & Senoufo, Tagwana                   & wal                             & Wolaytta                           \\
pan                             & Eastern Panjabi                    & sin                             & Sinhala                            & tha                             & Thai                               & war                             & Waray                              \\
pap                             & Papiamento                         & slk                             & Slovak                             & thk                             & Tharaka                            & wal                             & Wolaytta                           \\
pbt                             & Southern Pashto                    & slv                             & Slovenian                          & thv                             & Tamahaq, Tahaggart                 & wbi                             & Vwanji                             \\
pcm                             & Nigerian Pidgin                    & smo                             & Samoan                             & tke                             & Takwane                            & wec                             & Guere                              \\
pem                             & Kipende                            & sna                             & Shona                              & tir                             & Tigrinya                           & wes                             & Pidgin, Cameroon                   \\
pes                             & Western Persian                    & snd                             & Sindhi                             & tiv                             & Tiv                                & wib                             & Toussian, Southern                 \\
pir                             & Piratapuyo                         & snf                             & Noon                               & tlj                             & Talinga-Bwisi                      & wmw                             & Mwani                              \\
pkb                             & Kipfokomo / Pokomo                 & sng                             & Sanga / Kiluba                     & tll                             & Otetela                            & wol                             & Wolof                              \\
plt                             & Plateau Malagasy                   & snw                             & Selee                              & tog                             & Tonga                              & won                             & Wongo                              \\
pol                             & Polish                             & som                             & Somali                             & toh                             & Gitonga                            & xan                             & Xamtanga                           \\
por                             & Portuguese                         & sop                             & Kisonge                            & toi                             & Chitonga                           & xed                             & Hdi                                \\
pov                             & Guinea-Bissau Creole               & sor                             & Somrai                             & tpi                             & Tok Pisin                          & xho                             & Isixhosa                           \\
poy                             & Pogolo / Shipogoro-Pogolo          & sot                             & Sesotho                            & tpm                             & Tampulma                           & xnz                             & Mattokki                           \\
quy                             & Ayacucho Quechua                   & soy                             & Miyobe                             & tsc                             & Tshwa                              & xog                             & Soga                               \\
rag                             & Lulogooli                          & spa                             & Spanish                            & tsn                             & Setswana                           & xon                             & Konkomba                           \\
rel                             & Rendille                           & spp                             & Senoufo, Supyire                   & tso                             & Tsonga                             & xpe                             & Kpelle                             \\
rif                             & Tarifit                            & srd                             & Sardinian                          & tsw                             & Tsishingini                        & xrb                             & Karaboro, Eastern                  \\
rim                             & Nyaturu                            & srp                             & Serbian                            & ttj                             & Toro / Rutoro                      & xsm                             & Kasem                              \\
rnd                             & Uruund                             & ssw                             & Siswati                            & ttq                             & Tawallammat                        & xtc                             & Katcha-Kadugli-Miri                \\
rng                             & Ronga / ShiRonga                   & suk                             & Sukuma                             & ttr                             & Nyimatli                           & xuo                             & Kuo                                \\
ron                             & Romanian                           & sun                             & Sundanese                          & tui                             & Toupouri                           & yal                             & Yalunka                            \\
rub                             & Gungu                              & sus                             & Sosoxui                            & tuk                             & Turkmen                            & yam                             & Yamba                              \\
run                             & Rundi / Kirundi                    & swa                             & Swahili                            & tul                             & Kutule                             & yao                             & Yao / Chiyao                       \\
rus                             & Russian                            & swc                             & Swahili Congo                      & tum                             & Chitumbuka                         & yat                             & Yambeta                            \\
rwk                             & Rwa                                & swe                             & Swedish                            & tur                             & Turkish                            & yba                             & Yala                               \\
sag                             & Sango                              & swh                             & Swahili                            & tuv                             & Turkana                            & ybb                             & Yemba                              \\
saq                             & Samburu                            & swa                             & Swahili                            & tvu                             & Tunen                              & ydd                             & Eastern Yiddish                    \\
san                             & Sanskrit                           & swc                             & Swahili Congo                      & twi                             & Twi                                & yom                             & Ibinda                             \\
sat                             & Santali                            & swe                             & Swedish                            & uig                             & Uyghur                             & yor                             & Yoruba                             \\
sba                             & Ngambay                            & swh                             & Swahili                            & ukr                             & Ukrainian                          & yre                             & Yaoure                             \\
sbd                             & Samo, Southern                     & swk                             & Sena, Malawi                       & umb                             & Umbundu                            & yue                             & Yue Chinese                        \\
sbp                             & Sangu                              & sxb                             & Suba                               & urd                             & Urdu                               & zaj                             & Zaramo                             \\
sbs                             & Kuhane                             & szl                             & Silesian                           & urh                             & Urhobo                             & zdj                             & Comorian, Ngazidja                 \\
sby                             & Soli                               & tam                             & Tamil                              & uth                             & ut-Hun                             & zga                             & Kinga                              \\
scn                             & Sicilian                           & taq                             & Tamasheq                           & uzn                             & Northern Uzbek                     & zho                             & Chinese (Simplified)               \\
sef                             & Sénoufo, Cebaara                   & tat                             & Tatar                              & vag                             & Vagla                              & ziw                             & Zigula                             \\
ses                             & Songhay, Koyraboro Senni           & tel                             & Telugu                             & vai                             & Vai                                & zne                             & Zande / paZande                    \\
sev                             & Sénoufo, Nyarafolo                 & tcc                             & Datooga                            & vec                             & Venetian                           & zsm                             & Standard Malay                     \\
sfw                             & Sehwi                              & tcd                             & Tafi                               & ven                             & Tshivenda                          & zul                             & Isizulu                            \\
sgw                             & Sebat Bet Gurage                   & ted                             & Krumen, Tepo                       & vid                             & Chividunda                         &                                 &                                    \\
shi                             & Tachelhit                          & tem                             & Timne                              & vie                             & Vietnamese                         &                                 &                                    \\
shj                             & Shatt                              & tel                             & Telugu                             & vif                             & Vili                               &                                 &                                   
\\
\bottomrule
\end{tabular}%
}

\caption{List of languages in Babel-670 - Part III.}
\label{tab:lang_listIII}
\end{table*}

\end{document}